\documentclass[review]{elsarticle}

\usepackage[colorlinks,urlcolor=blue,linkcolor=blue,citecolor=blue]{hyperref}
\usepackage{graphicx}
\usepackage{amsmath,amssymb,amsfonts}
\usepackage{bm}
\usepackage{algpseudocode}
\usepackage{booktabs}
\usepackage{multirow}
\usepackage{array}
\usepackage{caption}
\usepackage{rotating}
\usepackage{adjustbox}
\usepackage{tabularx}
\usepackage{url}
\usepackage{float}
\usepackage{natbib}
\usepackage{xcolor}
\usepackage{tikz}
\usetikzlibrary{arrows.meta,positioning,calc,decorations.pathreplacing}

\definecolor{encblue}{HTML}{0072B2}
\definecolor{decorange}{HTML}{E69F00}
\definecolor{softgreen}{HTML}{009E73}
\definecolor{hardred}{HTML}{D55E00}
\definecolor{obsgray}{HTML}{999999}
\definecolor{tgtpurple}{HTML}{CC79A7}

\newlength{\tblwidth}
\newcolumntype{L}{>{\raggedright\arraybackslash}X}
\newcolumntype{C}{>{\centering\arraybackslash}X}
\newcolumntype{R}{>{\raggedleft\arraybackslash}X}

\biboptions{authoryear}

\numberwithin{theorem}{section}
\numberwithin{lemma}{section}

\usepackage{etoolbox}
\makeatletter
\apptocmd{\@floatboxreset}{\small}{}{}
\makeatother

\providecommand{\safeincludegraphics}[2][]{\includegraphics[#1]{#2}}

\begin{document}

\begin{frontmatter}

\title{Neuro-Symbolic Hierarchical Intention Anticipation in Human Behavior}

\author[a]{Farnaz Soleimani}
\ead{farnaz.soleimani@u-pec.fr}

\author[a]{Abdelghani Chibani}
\ead{abdelghani.chibani@u-pec.fr}

\author[a]{Yacine Amirat}
\ead{yacine.amirat@u-pec.fr}

\author[a]{Ghazaleh Khodabandelou\corref{cor1}}
\ead{ghazaleh.khodabandelou@u-pec.fr}
\cortext[cor1]{Corresponding author.}

\address[a]{LISSI Laboratory, University of Paris-Est Créteil (UPEC), IUT de Créteil-Vitry, France}
\begin{abstract}
Assistive autonomous systems must anticipate human goals before an observed behavior is complete. This article formulates anticipation as goal inference from a partially observed multimodal episode together with structured prediction of the remaining behavior, rather than exact motor forecasting. A compact Hierarchical Planning Decoder (HPD) is attached to a frozen neuro-symbolic recognition encoder and predicts, at four ontological levels, the next actions, the remaining activities and low-level intentions, and the episode high-level intention (HLI). The decoder is trained with soft neuro-symbolic regularization combining transition-coherence and hierarchical-continuity losses, and is decoded with hard reachability masks that enforce ontological validity at inference. On a compositional four-level benchmark of $15{,}002$ multimodal episodes built over NTU RGB+D 120 features, three headline properties are observed together. The advantage over the strongest sequential baseline grows with the anticipation horizon, from $+1.7$ points at step 1 to $+7.3$ points at step 3 (top-5). Under compositional generalization, where one parent association per multi-parent low-level intention is held out, this advantage widens to $+4.9$ points at step 1. At the episode level, $96.8\%$ of anticipated trajectories satisfy the joint logic constraints, above the $88.1\%$ strongest-baseline value and the $73.9\%$ ground-truth floor; soft logic terms alone account for a $59.8$ to $71.1\%$ relative reduction of HLI-reachability violations, and the hard masks then eliminate them entirely. Neural generation supplies predictive ranking, symbolic constraints supply ontological validity, and their combination yields coherent hierarchical anticipation while exposing remaining challenges in compositional goal generalization and unordered set prediction.
\end{abstract}


\end{frontmatter}

\section{Introduction}
\label{sec:introduction}
\subsection{Motivation and Context}
\label{ssec:motivation}

Reliable human-centric autonomous systems, ranging from assistive robots in domestic environments to collaborative agents in industrial workspaces, depend on the ability to act \emph{before} a person's behavior completes. Recognizing what has already happened is no longer sufficient; the system must infer the goal that drives the ongoing behavior and predict how the remainder of the behavior is likely to unfold, so that assistance can be planned rather than improvised. 
{ This shift from reactive recognition to proactive anticipation is a recurring theme in recent work on human-robot collaboration \cite{li2023proactive}, and is reflected in the rapid expansion of anticipation benchmarks \cite{damen2022rescaling, grauman2022ego4d, perrett2025hd}.}

The companion recognition work \cite{soleimani:hal-05616609} established a neuro-symbolic foundation for hierarchical \emph{recognition}. A Neuro-Symbolic Graph Transformer (NSGT), augmented with a learnable Human Reasoning Graph (HR-G) attention bias and a differentiable First-Order Logic (FOL) regulariser, was shown to jointly classify atomic actions, composite activities, Low-Level Intentions (LLI), and High-Level Intentions (HLI), improving logic consistency with only a small trade-off in action macro-F1. The present work asks whether the same neuro-symbolic foundation can be extended from \emph{describing the present} to \emph{anticipating the future}.

The delta relative to the recognition companion is precise. Recognition operated on complete episodes and produced classification decisions for the four ontological levels; it neither observed a prefix nor produced a forecast. The present work addresses partial-observation goal inference from a truncated prefix, generative prediction of the ordered remaining actions with end-of-sequence, prediction of the two remaining-set targets, and hard ontological guarantees on the entire generated trajectory. None of these was part of the recognition setting.

The conceptualization of anticipation on hierarchical behavior data requires methodological clarification. This concept is shaped by two properties of the utilized evaluation benchmark \cite{soleimani2026benchmark}, necessitating a graded rather than absolute interpretation. When an order-destroying control is applied, the sensitivity of episode-level \emph{classification} to sibling order is observed to decrease with the hierarchy level, ultimately falling within seed variation at the intention levels. Consequently, a goal is essentially recoverable from an unordered composition of its constituents. Concurrently, a strong \emph{step-wise} first-order signal at every level is demonstrated by a training-free predictability audit. It is shown that conditioning on the previous item increases the next-item top-5 accuracy by $+38.7$ percentage points at the action level, $+23.2$ at the activity level, and $+31.3$ at the LLI level over the marginal predictor. Therefore, anticipation is improved by pairing sequential next-action prediction, where local first-order structure is strong, with set-valued and episode-level targets, where such structure is absent. The design of the transition-coherence loss is built upon this reconciliation (Section~\ref{sec:methodology}). Anticipation is therefore formulated in this work as \emph{goal inference from a partially observed episode combined with prediction of the remaining hierarchical behavior}, rather than as precise motor-level forecasting of the exact next movement. Concretely, given a prefix covering a fraction $\rho$ of an episode, the system infers the episode's HLI, predicts the set of remaining activities and LLIs, and proposes next actions that remain coherent with the benchmark's released transition model.

A growing body of work has tackled action anticipation as an isolated forecasting task. Early recurrent approaches \cite{furnari2020rollingulstm, osman2021slowfast} were progressively replaced by transformer-based aggregators \cite{girdhar2021anticipative, zhong2023anticipative, diko2024semanticallygear}, and most recently by large language model (LLM) decoders that exploit commonsense priors over action sequences \cite{kim2023palm, qiu2025lta2025, cao2025insight}. Despite considerable progress, three structural limitations persist. First, anticipated outputs are typically restricted to a flat label space; the intermediate cognitive layers between observed motion and inferred long-term goals are either absent or collapsed into a single latent scenario variable \cite{mascaro2023intention}. Second, LLM-based decoders produce sequences that are linguistically fluent but logically unconstrained, yielding forecasts that contradict ontological axioms, for example anticipating a sequence that simultaneously belongs to mutually exclusive long-term intentions.

The framework of the companion recognition work is extended with a Hierarchical Planning Decoder (HPD) and two new categories of logical constraint, reformulating anticipation as structured generative reasoning. Future behavior is not treated as independent label draws but as an ontologically grounded trajectory. The training objective combines the anticipation task losses with two differentiable logic terms, a transition-coherence term and a hierarchical-continuity term, and this soft objective is complemented at inference by hard reachability constraints. The result encourages an anticipated trajectory that is consistent with the transition structure of the behavior model and with the high-level intention inferred from observed evidence, and it guarantees ontological reachability at decoding time.

\subsection{Problem Statement and Contributions}
\label{ssec:contributions}

The paper addresses two methodological bottlenecks in current anticipation systems.

\paragraph{- \textbf{Hierarchical anticipation}}
Most anticipation models predict future actions in a flat label space. Even intention-conditioned methods \cite{mascaro2023intention, cao2025insight} usually represent intention as a single latent scenario rather than as a structured ontology. This limits both interpretability and long-horizon reasoning: an atomic action can be compatible with several activities, and an activity can support different low- and high-level intentions depending on the surrounding context.

\paragraph{- \textbf{Ontologically constrained generation}}
Modern sequence decoders, including transformer and LLM-based anticipators, can generate plausible action continuations while violating the semantic constraints that define valid behavior trajectories. In assistive settings, such inconsistencies are not cosmetic: a system that predicts mutually incompatible goals cannot support reliable proactive assistance. The benchmark used here provides typed FOL rules as evaluation instruments; this work integrates them into both training and inference.

{ The contributions are:
\begin{itemize}
\item \textbf{A hierarchical anticipation formulation.} Anticipation is posed as partial-observation goal inference plus prediction of the remaining behavior at four levels: future actions, remaining activities, remaining LLIs, and HLI.

\item \textbf{A compact Hierarchical Planning Decoder.} A $2.06$M-parameter autoregressive HPD is attached to a frozen neuro-symbolic recognition encoder. The decoder receives the same observed action labels as the strongest label-oracle baselines, making the comparison protocol explicit.

\item \textbf{A soft/hard neuro-symbolic mechanism.} Differentiable Type-D transition-coherence and Type-E hierarchical-continuity losses guide training, while inference-time reachability masks enforce hard ontology constraints on the action and set outputs. The soft terms alone account for a $59.8$ to $71.1\%$ relative reduction of HLI-reachability violations before any decoding constraint is applied.

\item \textbf{Horizon-scaled and compositional gains under coherence guarantees.} On $15{,}002$ multimodal episodes, the advantage of the final system over the strongest sequential baseline grows from $+1.7$ points at step 1 to $+7.3$ points at step 3 (top-5), and widens to $+4.9$ points at step 1 on the held-out compositional split. Episode-level satisfaction of the joint logic constraints reaches $96.8\%$, above the $88.1\%$ best-baseline value and above the $73.9\%$ ground-truth floor.
\end{itemize}}

The remainder of the article is organized as follows. Section~\ref{sec:related_work} reviews short-term anticipation, long-term intention-conditioned forecasting, and neuro-symbolic sequence reasoning. Section~\ref{sec:methodology} describes the frozen recognition encoder, the HPD, and the soft/hard FOL mechanism. Section~\ref{sec:experimental_setup} details the benchmark, splits, baselines, and metrics. Section~\ref{sec:results_j2} presents the quantitative results and ablations. Section~\ref{sec:discussion_j2} discusses the findings and limitations. Section~\ref{sec:conclusion_j2} concludes.

\section{Related Work}
\label{sec:related_work}

The proposed framework intersects three lines of research: short-term action anticipation, long-term and intention-conditioned forecasting, and neuro-symbolic approaches to sequence reasoning. Each is reviewed in turn, after which the positioning of the present work is articulated.

\subsection{Short-Term Action Anticipation}
\label{ssec:short_term_anticipation}

Short-term action anticipation aims at predicting the next action a person will perform $\tau_a$ seconds before its onset, conditioned on a video segment of observed length $\tau_o$. The task was formalized on EPIC-KITCHENS \cite{damen2018scaling, damen2022rescaling} and has since become the standard formulation for first-person predictive video understanding.

Early architectures relied on recurrent modeling of past observations. Rolling-Unrolling LSTMs \cite{furnari2020rollingulstm} decomposed the problem into a summarization stage and a forecasting stage, fusing RGB, optical flow, and object-based features through a learned modality-attention mechanism. Multi-modal temporal convolutional networks \cite{osman2021slowfast} subsequently replaced recurrence with hierarchical dilated convolutions to reduce inference latency.

The introduction of self-attention shifted the field toward transformer-based aggregators. The Anticipative Video Transformer \cite{girdhar2021anticipative} computed causal attention across observed frames to forecast future actions, and the Anticipative Feature Fusion Transformer \cite{zhong2023anticipative} extended this idea to multi-modal token streams. More recent methods inject auxiliary semantic structure: S-GEAR \cite{diko2024semanticallygear} aligns visual prototypes with the geometry of action label semantics, and Action-Guided Attention \cite{tai2026action} replaces dot-product attention with a representation explicitly informed by recognized past actions. Label-smoothing strategies \cite{camporese2021knowledge} and text-based modalities \cite{ghosh2023text} have further been shown to improve generalization.

A common limitation of these approaches is that they operate on a flat output space and do not model the cognitive hierarchy that connects atomic actions to longer-term goals. Moreover, they implicitly assume that the exact next motor action is well-determined by the observed prefix. On compositional behavior, in which a goal admits many valid orderings of its parts, this assumption breaks down; the present work adopts a goal-centric formulation instead.

\subsection{Long-Term Anticipation and Intention-Conditioned Forecasting}
\label{ssec:long_term_anticipation}

The Long-Term Action Anticipation (LTA) benchmark introduced with Ego4D \cite{grauman2024ego4d} requires the prediction of up to twenty future actions from extended observations, shifting emphasis from immediate motor prediction toward goal-directed reasoning. The Ego4D baseline pairs a SlowFast \cite{feichtenhofer2019slowfast} visual encoder with a transformer aggregator and parallel classification heads; this design is now standard.

Intention-conditioned variants explicitly factorize the prediction into a goal-inference stage followed by an action-generation stage. ICVAE \cite{mascaro2023intention} introduced a scenario variable, interpretable as the camera wearer's intention, that conditions a transformer decoder and won the CVPR/ECCV 2022 Ego4D LTA challenges. This established that explicit intention modeling, rather than implicit context aggregation, improves long-horizon forecasting.

Subsequent work has leaned heavily on large language models as the reasoning backbone. PALM \cite{kim2023palm} chains a captioning module and an LLM to anticipate future action sequences from textual descriptions of past actions, demonstrating that commonsense priors encoded in pretrained LLMs transfer to LTA without task-specific training. QueryMamba \cite{zhong2024querymamba} replaces the transformer decoder with a Mamba state-space model and adds a statistical co-occurrence module. The 2025 Ego4D LTA challenge winner \cite{qiu2025lta2025} combines a high-capacity visual encoder, a transformer recognition module, and a fine-tuned LLM, formalizing a three-stage pipeline that has become a de facto template. Vision-and-intention LLM approaches \cite{cao2025vision} jointly condition the language model on scene observations and inferred goals, while INSIGHT \cite{cao2025insight} integrates a reinforcement learning objective with a structured reasoning template. HD-EPIC \cite{perrett2025hd} extends EPIC-KITCHENS with dense annotation, enabling finer-grained anticipation.

Across this line of work, two limitations recur. First, intention is typically treated as a single latent variable or scenario label rather than as a structured multi-level hierarchy with explicit ontological semantics. Second, when generative decoders are employed, the output space is unconstrained by symbolic axioms, allowing sequences that are fluent but ontologically inconsistent with the observed context.

\subsection{Neuro-Symbolic Reasoning for Sequence Forecasting and Planning}
\label{ssec:neurosymbolic_forecasting}

Neuro-symbolic AI \cite{hitzler2022neuro, garcez2023neurosymbolic} combines the perceptual capacity of neural networks with the structural soundness of declarative reasoning. While early integrations such as DeepProbLog \cite{manhaeve2019deepproblog} and Logic Tensor Networks \cite{badreddine2022logic} have demonstrated the value of differentiable logic on static tasks, their extension to temporally extended forecasting is more recent.

In the context of action anticipation, Bhagat et al.\ \cite{bhagat2023knowledge} augment a transformer's attention mechanism with a symbolic knowledge graph encoding object affordances, reporting up to a nine-percentage-point improvement over purely neural baselines on the Breakfast and 50 Salads datasets. The follow-up NeSCA framework \cite{bhagat2024let} extends this design to short-context anticipation in collaborative cooking, demonstrating that symbolic priors halve the observation window required for accurate prediction. Bellotto et al.\ \cite{mghames2023neuro} apply a related strategy to human motion prediction.

Beyond anticipation specifically, hierarchical neuro-symbolic architectures have emerged for general planning. The Hierarchical Neuro-Symbolic Decision Transformer \cite{baheri2025hierarchical} couples a classical symbolic planner with a transformer-based low-level policy via a bidirectional interface, separating logically coherent operator sequencing from reactive control. HVR \cite{cornelio2025hierarchical} integrates hierarchical task decomposition, retrieval-augmented generation over a symbolic knowledge graph, and a symbolic validator that simulates plans before execution. NeSyA \cite{manginas2024nesya} integrates neural perception with symbolic automata for sequence classification.

These works confirm that explicit symbolic structure can be made differentiable and that it improves both accuracy and explainability in temporal reasoning. They do not, however, provide a unified treatment of (i) multimodal fusion, (ii) a deep ontological hierarchy spanning four levels of abstraction, and (iii) anticipatory generation under logical constraints. The companion recognition work \cite{soleimani:hal-05616609} provides the first two; the present work extends the framework to the third.

\subsection{Positioning of the Present Work}
\label{ssec:positioning_j2}

The literature leaves a precise gap. Existing anticipation systems are strong at flat next-action or long-term sequence forecasting, but they do not jointly anticipate atomic actions, activities, LLIs, and HLIs under an explicit ontology. Neuro-symbolic methods have introduced logical structure into recognition, attention, or planning, but their use as a training-time and inference-time constraint on generated anticipation trajectories remains limited. This work therefore positions itself between action anticipation and neuro-symbolic planning: the decoder remains data-driven, but the generated future is required to remain reachable under the task ontology. The comparison with a symbolic-only OntoPrior, a neural-only HPD, and the final soft/hard HPD is designed to test exactly this claim.

\section{Methodology}
\label{sec:methodology}

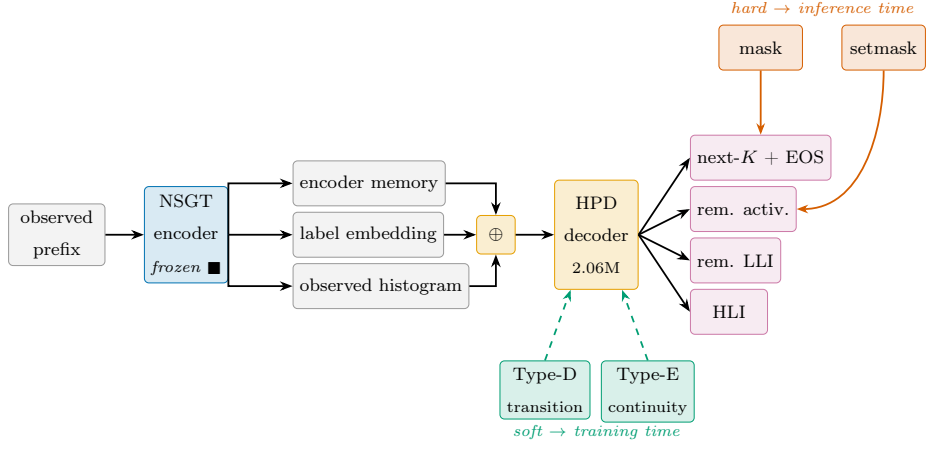
\begin{figure}[t]
\centering
\resizebox{\columnwidth}{!}{%
\begin{tikzpicture}[
  font=\footnotesize,
  box/.style={draw,rounded corners=2pt,minimum height=7mm,minimum width=13mm,align=center,inner sep=3pt},
  enc/.style={box,fill=encblue!15,draw=encblue,line width=0.4pt},
  dec/.style={box,fill=decorange!18,draw=decorange,line width=0.4pt},
  soft/.style={box,fill=softgreen!15,draw=softgreen,line width=0.4pt},
  hard/.style={box,fill=hardred!15,draw=hardred,line width=0.4pt},
  inp/.style={box,fill=obsgray!12,draw=obsgray,line width=0.4pt,minimum width=15mm},
  outp/.style={box,fill=tgtpurple!15,draw=tgtpurple,line width=0.4pt,minimum width=12mm},
  flow/.style={-{Stealth[length=2mm]},thick},
  softflow/.style={-{Stealth[length=1.8mm]},thick,softgreen,dashed},
  hardflow/.style={-{Stealth[length=1.8mm]},thick,hardred},
]
\node[inp] (prefix) {observed\\prefix};
\node[enc,right=6mm of prefix,minimum height=15mm] (encoder)
  {NSGT\\encoder\\{\scriptsize\itshape frozen \raisebox{-0.3ex}{$\blacksquare$}}};
\node[inp,right=10mm of encoder,yshift=8mm] (mem) {encoder memory};
\node[inp,right=10mm of encoder,yshift=0mm] (lab) {label embedding};
\node[inp,right=10mm of encoder,yshift=-8mm] (hist) {observed histogram};
\node[dec,right=5mm of lab,minimum height=6mm,minimum width=6mm] (fuse) {$\oplus$};
\node[dec,right=6mm of fuse,minimum height=17mm] (hpd) {HPD\\decoder\\{\scriptsize 2.06M}};
\node[outp,right=8mm of hpd,yshift=12mm] (onext) {next-$K$ + EOS};
\node[outp,right=8mm of hpd,yshift=4mm] (oact) {rem. activ.};
\node[outp,right=8mm of hpd,yshift=-4mm] (olli) {rem. LLI};
\node[outp,right=8mm of hpd,yshift=-12mm] (ohli) {HLI};
\node[soft,below=11mm of hpd,xshift=-8mm] (typed) {Type-D\\{\scriptsize transition}};
\node[soft,below=11mm of hpd,xshift=8mm] (typee) {Type-E\\{\scriptsize continuity}};
\node[hard,above=10mm of onext] (mask) {mask};
\node[hard,right=6mm of mask] (setmask) {setmask};
\draw[flow] (prefix) -- (encoder);
\draw[flow] (encoder.east) |- (mem.west);
\draw[flow] (encoder.east) -- (lab.west);
\draw[flow] (encoder.east) |- (hist.west);
\draw[flow] (mem.east) -| (fuse.north);
\draw[flow] (lab.east) -- (fuse.west);
\draw[flow] (hist.east) -| (fuse.south);
\draw[flow] (fuse) -- (hpd);
\draw[flow] (hpd.east) -- (onext.west);
\draw[flow] (hpd.east) -- (oact.west);
\draw[flow] (hpd.east) -- (olli.west);
\draw[flow] (hpd.east) -- (ohli.west);
\draw[softflow] (typed.north) -- ([xshift=-4mm]hpd.south);
\draw[softflow] (typee.north) -- ([xshift=4mm]hpd.south);
\draw[hardflow] (mask.south) -- (onext.north);
\draw[hardflow] (setmask.south) to[out=-90,in=0] (oact.east);
\node[softgreen,font=\scriptsize\itshape] at ($(typed)!0.5!(typee)+(0,-6.5mm)$) {soft $\rightarrow$ training time};
\node[hardred,font=\scriptsize\itshape] at ($(mask)!0.5!(setmask)+(0,6mm)$) {hard $\rightarrow$ inference time};
\end{tikzpicture}%
}
\caption{The frozen graph NSGT encoder produces per-clip memory and a pooled HLI state from the observed prefix. The HPD fuses three prefix views, encoder memory, observed-label embeddings, and the observed-action histogram, then decodes the next actions with EOS, the remaining activity and LLI sets, and the episode HLI. Green dashed paths are the soft Type-D and Type-E logic losses applied during training; red solid blocks are the hard reachability masks applied at inference (mask on the action stream, setmask on the set heads).}
\label{fig:j2_pipeline}
\end{figure}

The recognition backbone of the companion work \cite{soleimani:hal-05616609} is reused as a frozen encoder, and a Hierarchical Planning Decoder (HPD) is trained on top of its prefix representations. The method separates learning from constraint enforcement. During training, differentiable logic losses bias the decoder toward coherent futures. During inference, reachability masks enforce hard ontology constraints without modifying the learned weights. Figure~\ref{fig:j2_pipeline} provides an overview.

\subsection{Frozen Recognition Encoder}
\label{ssec:frozen_backbone}

The encoder is the graph NSGT of the companion work: a graph-structured hierarchical transformer over the typed episode graph, operating on the four fused modalities with $d_{\mathrm{model}} = 256$, three graph layers, four heads, and cross-attention fusion. By pre-specified protocol, the checkpoint used for all anticipation experiments is the project default seed configuration (FOL condition \texttt{all}, seed 42); this checkpoint is fixed in advance rather than selected on anticipation scores, which avoids best-seed selection bias. The encoder holds $6.75$M parameters and is frozen throughout: a reproducibility check verifies its full-observation recognition HLI top-1 of $69.7\%$ to within $10^{-6}$ before every anticipation run, guaranteeing that the encoder contributes an unchanging representation.

For an observed prefix, the frozen encoder exposes two products consumed by the decoder: a sequence of per-clip encoder states, one per observed action clip, and a pooled high-level state that summarises the prefix at the HLI level. Neither the encoder weights nor these products change during decoder training.

The decision to freeze the encoder was verified empirically. Two fine-tuning regimes were evaluated with three seeds each, in every case with the extended FOL loss and constrained decoding kept identical to the main configuration and the encoder learning rate set to one tenth of the decoder's ($3\times10^{-5}$). Regime R1 unfroze the last HGT layer together with the cross-attention fusion, yielding $49$ unfrozen tensors alongside the HPD; regime R2 unfroze the full encoder for $152$ unfrozen tensors. Neither regime produced gains outside seed variance: the best case, R2 on the test split, moved step-1 top-1 by $+0.5$ points and step-1 top-5 by $+0.4$ points relative to the frozen encoder (Table~\ref{tab:appendix_std}), while HLI top-1 moved by $-0.4$ points; on the compositional split the deltas were within $\pm 0.6$ points across every metric. E1 and E2 remained exactly zero in every fine-tuned condition, again confirming that the reachability guarantee is structural. As a sanity check, a full-episode recognition probe on the fine-tuned encoders showed HLI top-1 changes between $-1.2$ and $+0.5$ points relative to the $69.7\%$ reference, so recognition competence is preserved throughout. The frozen configuration is therefore retained as the main system because it delivers the same anticipation quality at lower training cost, preserves the clean attribution across architecture, features, and logic that freezing makes possible, and leaves the encoder available for other downstream tasks without re-adaptation.
\subsection{Hierarchical Planning Decoder}
\label{ssec:planning_decoder}

The HPD is a small autoregressive transformer decoder with $d_{\mathrm{model}} = 192$, three layers, four heads, and a feedforward width of $768$, totalling $2.06$M trainable parameters. It is deliberately kept an order of magnitude smaller than the encoder so that the anticipation results are attributable to the reasoning design rather than to decoder capacity.

\paragraph{\textbf{Task-legal input}}
By pre-specified protocol, the decoder receives the observed action labels as input, in addition to the frozen encoder states.
This is a deliberate protocol-parity choice with the strongest baselines, which are label-oracle models that also see the observed labels; supplying the same labels to the HPD makes the comparison fair rather than advantageous. The choice does not commit the deployed system to relying on oracle labels: Section~\ref{ssec:results_robustness} evaluates the same trained decoder when the observed labels are replaced by the recognition encoder's predictions and under synthetic label-noise sweeps, and reports a noise-aware training variant that closes most of the gap to the oracle condition. Concretely, each observed position fuses its encoder state with an embedding of its action label, the embedding matrix being shared with the decoder output layer.
Formally, the fused state at observed position $i$ is
\begin{equation}
\label{eq:am1_fusion}
\mathbf{m}_i \;=\; W_{\mathrm{mem}}\,\mathbf{e}_i \;+\; W_{\mathrm{lab}}\,\mathrm{embed}(a_i),
\end{equation}
where $\mathbf{e}_i$ is the frozen encoder state, $a_i$ is the observed action label, and $\mathrm{embed}(\cdot)$ is the embedding shared with the decoder output layer.

A projection of the $85$-dimensional histogram of observed action counts is concatenated with the mean fused state and the pooled HLI state, and the result is projected to the decoder width to form the conditioning summary. This makes explicit that goal inference may draw on three distinct views of the prefix: the ordered encoder states, the ordered observed labels, and the unordered action histogram.

This protocol should not be interpreted as raw video-only anticipation. The HPD and the strongest baselines receive observed action labels, making the setting a structured anticipation problem conditioned on recognized or oracle-observed prefix labels. This choice isolates the contribution of hierarchical decoding and neuro-symbolic constraints from the upstream action-recognition problem.

\paragraph{\textbf{Outputs}}
Four heads read the decoder. An autoregressive next-action head emits up to $K = 3$ future actions over a vocabulary extended with an end-of-sequence token, terminating early through that token. Three further heads, reading the pooled decoder state, produce the remaining-activity set, the remaining-LLI set, and the episode HLI.

\paragraph{\textbf{Optional language-model variant}}
A separate ablation replaces the trained decoder with a LoRA-adapted small open-weight language model that emits the full structured target as text. It is documented in Section~\ref{ssec:results_llm} and is not part of the primary system.
\subsection{Ontological Constraint Semantics}
\label{ssec:ontology_constraints}

Let $\mathcal{A}$, $\mathcal{V}$, $\mathcal{L}$, and $\mathcal{H}$ denote the finite and pairwise disjoint sets of actions, activities, low-level intentions (LLIs), and high-level intentions (HLIs). The hierarchical structure is defined by a mereological part-of relation
\[
\sqsubset \;\subseteq\; 
(\mathcal{A}\times\mathcal{V}) \cup
(\mathcal{V}\times\mathcal{L}) \cup
(\mathcal{L}\times\mathcal{H}),
\]
rather than by an is-a or subsumption relation. Thus, an action is not a subclass of an activity, and an activity is not a subclass of an intention; lower-level elements are constituents of higher-level behavioral units.

Let $\sqsubset^{+}$ denote the transitive closure of $\sqsubset$. For a predicted HLI $h\in\mathcal{H}$, the reachable set is
\[
R(h)=\{x \in \mathcal{A}\cup\mathcal{V}\cup\mathcal{L} \mid x\sqsubset^{+} h\},
\]
with type-specific restrictions
\[
R_{\mathcal{A}}(h)=R(h)\cap\mathcal{A},\quad
R_{\mathcal{V}}(h)=R(h)\cap\mathcal{V},\quad
R_{\mathcal{L}}(h)=R(h)\cap\mathcal{L}.
\]
These sets are precomputed offline as binary masks. The composition graph is a directed acyclic graph rather than a tree: some LLIs may belong to more than one HLI, so reachable sets can overlap. This multi-parent structure is the basis of the compositional split.

For an episode $e$, violations are defined on grounded predictions rather than on the ontology alone. We write $\mathrm{occ}(x,e)$ when an element $x$ occurs in episode $e$, $\mathrm{next}(a_i,a_j,e)$ when action instance $a_j$ immediately follows $a_i$, and $\mathrm{hasHLI}(e,h)$ when $e$ is assigned HLI $h$. The predicate $\mathrm{hasHLI}$ is functional:
\[
\forall e,\forall h_1,h_2\in\mathcal{H},\quad
\mathrm{hasHLI}(e,h_1)\wedge \mathrm{hasHLI}(e,h_2)
\Rightarrow h_1=h_2.
\]
This makes assigning a trajectory to mutually incompatible high-level intentions a well-defined failure.

Given a predicted HLI $\hat h$, predicted future actions $\hat{\mathbf a}=(\hat a_1,\ldots,\hat a_K)$, and a predicted remaining-activity set $\hat{\mathcal V}$, the reachability violations are audited as
\[
E1(e)=
\frac{1}{|\hat{\mathbf a}|}
\sum_{\hat a\in\hat{\mathbf a}}
\mathbb{1}\!\left[\hat a\notin R_{\mathcal A}(\hat h)\right],
\]
and
\[
E2(e)=
\frac{1}{|\hat{\mathcal V}|}
\sum_{\hat v\in\hat{\mathcal V}}
\mathbb{1}\!\left[\hat v\notin R_{\mathcal V}(\hat h)\right].
\]
E1 and E2 are treated as hard reachability constraints at inference time.

Concretely, for action logits $\ell_{\mathcal A}(a)$, the inference-time mask is
\[
\ell'_{\mathcal A}(a)=
\begin{cases}
\ell_{\mathcal A}(a), & a\in R_{\mathcal A}(\hat h)\cup\{\mathrm{EOS}\},\\
-\infty, & \text{otherwise}.
\end{cases}
\]
The same principle is applied to the remaining-activity and remaining-LLI set heads using $R_{\mathcal V}(\hat h)$ and $R_{\mathcal L}(\hat h)$.

By contrast, D1 and D2 are not hard ontological constraints. Let $\mathcal{S}_{\mathrm{train}}\subseteq\mathcal{A}\times\mathcal{A}$ be the successor support estimated from the training split. The transition-support audit is
\[
D1(e)=
\frac{1}{K-1}
\sum_{t=1}^{K-1}
\mathbb{1}\!\left[(\hat a_t,\hat a_{t+1})\notin\mathcal{S}_{\mathrm{train}}\right].
\]
This constraint is used only as a soft transition-coherence preference during training, because valid test trajectories may contain transitions unseen in the training split. D2, which concerns canonical within-activity ordering, is also diagnostic only, since the canonical order is defined only for a subset of activities and the ground truth itself may violate it.

When transition support and reachability disagree, reachability takes strict priority. Candidates outside $R(\hat h)$ are removed regardless of their transition score, whereas transition coherence only biases the ranking among candidates that remain ontologically admissible. Thus, E1 and E2 define hard validity, while D1 and D2 define softer plausibility or diagnostic criteria.

All sets and predicates are finite and fully enumerated. Once an episode is fixed, the constraint language reduces to a function-free ground fragment equivalent to a propositional theory over typed atoms. The neuro-symbolic component therefore does not perform open-domain first-order inference or description-logic reasoning; it uses differentiable relaxations of grounded constraints during training and precomputed masks during decoding.

\subsection{Extended First-Order Logic Loss (soft, training time)}
\label{ssec:extended_fol}

Building on the ontological semantics defined above, the benchmark provides two differentiable training-time relaxations: Type-D transition coherence and Type-E hierarchical continuity. They are soft constraints: they guide the probability distribution but do not guarantee validity by themselves.

\paragraph{\textbf{Type-D, transition coherence}}
Let $\mathcal{N}_{\mathrm{trans}}(a)$ denote the set of actions observed to follow action $a$ in the training episodes, and let $\mathbf{p}_{t+1}$ be the decoder distribution at future step $t+1$. After a decoded predecessor $\hat a_t$, transition coherence penalizes probability assigned outside the valid successor set:
\begin{equation}
\label{eq:fol_temporal}
\mathcal{L}_{\mathrm{D}}
= \frac{1}{K-1}\sum_{t=1}^{K-1}
\sum_{a' \notin \mathcal{N}_{\mathrm{trans}}(\hat a_t) \cup \{\mathrm{EOS}\}}
\mathbf{p}_{t+1}(a').
\end{equation}
This term is used as a coherence prior rather than as a claim that the next action is uniquely determined by temporal order. It is consistent with the benchmark design: episode-level labels are compositional and order-insensitive, while adjacent steps still carry local transition information.

\paragraph{\textbf{Type-E, hierarchical continuity}}
Let $\mathbf{p}_{H}^{\mathrm{obs}}$ be the HLI distribution inferred from the observed prefix and $\mathbf{p}_{H}^{\mathrm{antic}}$ the HLI distribution predicted by the HPD. Hierarchical continuity penalizes drift between the prefix-inferred goal and the anticipated goal:
\begin{equation}
\label{eq:fol_hierarchical}
\mathcal{L}_{\mathrm{E}}
= \mathrm{KL}\!\left(\mathbf{p}_{H}^{\mathrm{obs}}\,\|\,\mathbf{p}_{H}^{\mathrm{antic}}\right).
\end{equation}

The final training objective is
\begin{equation}
\label{eq:total_loss_j2}
\mathcal{L}
= \mathcal{L}_{\mathrm{CE}}^{\mathrm{antic}}
+ \lambda_{\mathrm{D}}\mathcal{L}_{\mathrm{D}}
+ \lambda_{\mathrm{E}}\mathcal{L}_{\mathrm{E}},
\end{equation}
with $\lambda_{\mathrm{D}}=\lambda_{\mathrm{E}}=0.5$. The task loss $\mathcal{L}_{\mathrm{CE}}^{\mathrm{antic}}$ comprises cross-entropy for the autoregressive next-action and HLI heads and binary cross-entropy for the remaining-activity and remaining-LLI set heads. The encoder is frozen, so no recognition loss is optimized during HPD training.

\subsection{Constrained Decoding (hard, inference time)}
\label{ssec:constrained_decoding}

At inference, the decoder first predicts the episode HLI. This predicted HLI defines the ontology subtree reachable by the anticipated trajectory. The action mask then adds $-\infty$ to the logits of every action that is not reachable from this HLI, while keeping the EOS token available. The same reachability principle is applied to the remaining-activity and remaining-LLI set heads through a set mask. These masks are not learned and do not change the decoder parameters; they only remove logically invalid outputs from the decoding space. The same trained decoder can therefore be evaluated with or without hard masking, which makes the contribution of soft training and hard inference constraints separable in the ablation study.

\subsection{Training Protocol}
\label{ssec:training_protocol}

Frozen encoder states are precomputed for every prefix. Only the $2.06$M HPD parameters are trained, using AdamW with learning rate $3\times10^{-4}$, batch size $256$, dropout $0.1$, and a maximum budget of $40$ epochs. The observation ratio is sampled during training so that one decoder can be evaluated at $r\in\{0.25,0.50,0.75\}$. Early stopping monitors validation step-1 top-1 through the same greedy decoding path used at test time. Learned configurations are trained with three seeds and reported as mean $\pm$ standard deviation. The best epoch occurs early (mean $7.2$, maximum $11$), indicating that the budget is not the limiting factor. The trained decoder is small ($2.06$M parameters) and decodes $K=3$ actions greedily with the three set and goal heads read in parallel, a configuration compatible with real-time serving in principle, though wall-clock latency under a target workload was not profiled here.

To keep the reported comparison auditable, all baselines and HPD variants are evaluated with the same record construction, vocabulary, split definitions, and metric code. The frozen encoder checkpoint is fixed before anticipation training, and its full-episode recognition score is checked before HPD training. This protocol is important because the anticipation task receives observed action labels, whereas the full-episode recognition reference does not; the two numbers are therefore not ceilings for one another.

\section{Experimental Setup}
\label{sec:experimental_setup}

\subsection{Benchmark}
\label{ssec:benchmark}

All experiments are conducted on a compositional four-level anticipation benchmark built over NTU RGB+D 120 features. The synthesis protocol originates in a companion benchmark manuscript \cite{soleimani2026benchmark}; the essential ontology, splits, features, and logic rules are summarized here to make the present article self-contained for methodological review.

\paragraph{\textbf{Ontology and episodes}}
Behavior is organized over four typed levels: an \emph{action} is an atomic, single-label clip from the source corpus; an \emph{activity} composes 1 to 2 actions; a \emph{low-level intention} (LLI) composes 2 to 4 activities; a \emph{high-level intention} (HLI) is a long-horizon episode of 2 to 5 LLIs. The instantiated ontology comprises $8$ HLIs, $35$ LLIs, $50$ activities, and $85$ source action classes. The benchmark contains $15{,}002$ episodes averaging $6.95$ actions each (median $6$, range $4$ to $17$), with the eight HLIs populated nearly uniformly. All clips within an episode are performed by the same subject, drawn from $60$ source performers under a coverage-aware sampler. Of the $85$ ontology action classes, $81$ occur in the synthesized episodes; the effective action vocabulary of every model in this article is therefore $81$ classes, and all reported action metrics are computed over this vocabulary.

\paragraph{\textbf{Features}}
Each action clip carries pooled per-clip features from four modalities: skeleton ($2048$-d), RGB ($1024$-d), infrared ($1024$-d), and depth ($1024$-d), concatenated into a fused $5120$-dimensional vector. Features are pre-extracted with frozen encoders and are identical across all models compared here.

\paragraph{\textbf{Splits}}
The subject-disjoint partition of the benchmark is reused unchanged: $7{,}062$ train, $1{,}519$ validation, $3{,}006$ test, and $3{,}415$ compositional-test episodes. The \emph{compositional test} split additionally withholds one parent association from training for each multi-parent LLI, so that at test time the concept must transfer to a high-level context never observed during training. Anticipation inherits this structure directly: every anticipation record is derived from exactly one episode of its split.

\paragraph{\textbf{Logic rules and intrinsic violation floors}}
The violation metrics follow the constraint semantics of Section~\ref{ssec:ontology_constraints}. We report four trajectory-level audits on completed predictions.

Four violation metrics are audited on completed trajectories. D1 flags a predicted next action whose (previous, next) pair falls outside the transition support observed in the training episodes. D2 flags a violation of the canonical within-activity ordering, defined for the $14$ of $50$ activities that declare one. E1 flags a predicted action that is not reachable from the predicted HLI through the ontology. E2 flags a predicted remaining activity that is not reachable from the predicted HLI. The ground-truth trajectories themselves establish intrinsic floors: D1 $= 0.064$ test and $0.007$ compositional test (the ground truth contains transitions unseen in train), D2 $= 0.224$ and $0.363$, and E1 $=$ E2 $= 0$ on both splits. Because its intrinsic floor is high, D2 is reported for completeness but excluded from all gating and headline claims. An episode-level satisfaction score is also reported: the fraction of records whose completed trajectory carries zero violations over $\{$D1, E1, E2$\}$; the ground-truth floor of this score is $0.739$ on test.

\subsection{Task Definition}
\label{ssec:task}

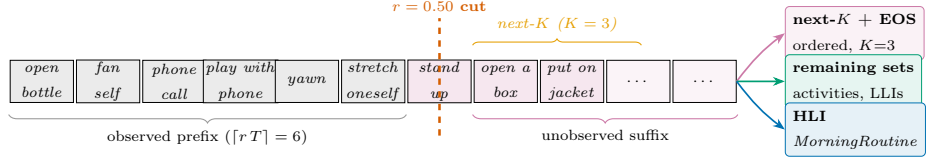
\begin{figure}[t]
\centering
\resizebox{\columnwidth}{!}{%
\begin{tikzpicture}[
  font=\footnotesize,
  obscell/.style={draw,fill=obsgray!20,minimum width=10.5mm,minimum height=7mm,inner sep=1pt,align=center,line width=0.4pt,font=\scriptsize},
  tgtcell/.style={draw,fill=tgtpurple!18,minimum width=10.5mm,minimum height=7mm,inner sep=1pt,align=center,line width=0.4pt,font=\scriptsize},
  fam/.style={draw,rounded corners=2pt,align=left,inner sep=3.5pt,font=\scriptsize,line width=0.4pt},
]
\node[obscell] (o1) at (0*1.10,0)  {\textit{open}\\\textit{bottle}};
\node[obscell] (o2) at (1*1.10,0)  {\textit{fan}\\\textit{self}};
\node[obscell] (o3) at (2*1.10,0)  {\textit{phone}\\\textit{call}};
\node[obscell] (o4) at (3*1.10,0)  {\textit{play with}\\\textit{phone}};
\node[obscell] (o5) at (4*1.10,0)  {\textit{yawn}};
\node[obscell] (o6) at (5*1.10,0)  {\textit{stretch}\\\textit{oneself}};
\node[tgtcell] (t1) at (6*1.10,0)  {\textit{stand}\\\textit{up}};
\node[tgtcell] (t2) at (7*1.10,0)  {\textit{open a}\\\textit{box}};
\node[tgtcell] (t3) at (8*1.10,0)  {\textit{put on}\\\textit{jacket}};
\node[tgtcell,fill=tgtpurple!8] (t4) at (9*1.10,0)  {$\cdots$};
\node[tgtcell,fill=tgtpurple!8] (t5) at (10*1.10,0) {$\cdots$};

\draw[hardred,very thick,dashed] (6.6,-0.9) -- (6.6,1.05);
\node[hardred,font=\scriptsize\bfseries,anchor=south] at (6.6,1.05) {$r=0.50$ cut};

\draw[decorate,decoration={brace,amplitude=4pt,mirror},obsgray,line width=0.5pt]
  (-0.55,-0.55) -- (6.05,-0.55) node[midway,below=4pt,font=\scriptsize,black] {observed prefix ($\lceil r\,T\rceil=6$)};
\draw[decorate,decoration={brace,amplitude=4pt,mirror},tgtpurple,line width=0.5pt]
  (7.15,-0.55) -- (11.55,-0.55) node[midway,below=4pt,font=\scriptsize,black] {unobserved suffix};

\draw[decorate,decoration={brace,amplitude=3pt},decorange,line width=0.5pt]
  (7.15,0.6) -- (9.95,0.6) node[midway,above=3pt,font=\scriptsize\itshape,decorange] {next-$K$ ($K=3$)};

\node[fam,fill=tgtpurple!10,draw=tgtpurple,right=8mm of t5,yshift=8mm] (fnext)
  {\textbf{next-$K$ + EOS}\\ordered, $K{=}3$};
\node[fam,fill=softgreen!10,draw=softgreen,right=8mm of t5,yshift=0mm] (fsets)
  {\textbf{remaining sets}\\activities, LLIs};
\node[fam,fill=encblue!10,draw=encblue,right=8mm of t5,yshift=-8mm] (fhli)
  {\textbf{HLI}\\\textit{MorningRoutine}};

\draw[-{Stealth[length=2mm]},thick,tgtpurple] (t5.east) to[bend left=8] (fnext.west);
\draw[-{Stealth[length=2mm]},thick,softgreen] (t5.east) -- (fsets.west);
\draw[-{Stealth[length=2mm]},thick,encblue] (t5.east) to[bend right=8] (fhli.west);
\end{tikzpicture}%
}
\caption{Task schematic instantiated on episode \texttt{ep\_0000331} (the same episode shown qualitatively in Fig.~\ref{fig:f8_qual}). The 11-action episode is cut at $r=0.50$, giving a 6-action observed prefix (\textit{open bottle}, \textit{fan self}, \textit{phone call}, \textit{play with phone}, \textit{yawn}, \textit{stretch oneself}) and a 5-action unobserved suffix. The three target families predicted from the prefix are: the ordered next-$K$ actions with EOS (here \textit{stand up}, \textit{open a box}, \textit{put on jacket}; EOS false since two further actions follow); the sets of remaining activities (\texttt{ACT\_Stand}, \texttt{ACT\_PackBag}, \texttt{ACT\_DonClothing}, \texttt{ACT\_DonAccessory}) and remaining LLIs (\texttt{LLI\_Awakening} in progress, \texttt{LLI\_DressUp}); and the episode HLI (\texttt{HLI\_MorningRoutine}).}
\label{fig:j2_task}
\end{figure}

Anticipation is instantiated as follows (Fig.~\ref{fig:j2_task}). Each episode of length $T$ actions is cut at an observation ratio $r$, giving an observed ordered prefix of $\lceil r\,T \rceil$ action clips (features, clip keys, and unit alignment retained). From the prefix, the system predicts: (i) the ordered next actions up to horizon $K = 3$, together with an end-of-sequence (EOS) flag when fewer than $K$ actions remain; (ii) the set of remaining activities; (iii) the set of remaining LLIs; and (iv) the episode HLI. The primary operating point is $r = 0.50$, with $r \in \{0.25, 0.75\}$ evaluated as sensitivity; no minimum-length filter is applied, so every episode of every split yields exactly one record per ratio. At $r = 0.50$, a mean of $3.19$ actions remains per episode and $62.9\%$ of episodes have at least $3$ remaining, so the $K = 3$ horizon is active for most records.

\paragraph{\textbf{Empirical justification of task well-posedness}}
Two measured properties of the data reconcile what may appear contradictory. On the one hand, the benchmark's order-destroying control shows that episode-level classification is insensitive to sibling order: shuffling changes recognition scores within seed variation. On the other hand, a training-free audit of next-step predictability shows a strong first-order signal at the step level: conditioning on the previous action lifts next-action top-5 accuracy by $+38.7$ percentage points over the marginal predictor and reduces the conditional entropy of the next action from $5.21$ to $3.51$ bits, with analogous lifts at the activity ($+23.2$) and LLI ($+31.3$) levels. Both facts are design properties, not contradictions: the episode label is determined by an unordered composition, while adjacent steps are generated by a transition model and therefore carry local order. The task accordingly pairs sequential next-action prediction, where local order is informative, with set-valued and episode-level targets, where it is not. A label-only logistic probe further confirms that goal inference from a prefix is feasible and improves with observation: HLI macro-F1 of $0.677$, $0.766$, and $0.821$ at $r = 0.25$, $0.50$, and $0.75$.

\paragraph{\textbf{Step-to-time convention}}

The benchmark carries no per-clip durations. Under the NTU RGB+D 120 convention, an action clip typically spans a few seconds; therefore, step-based horizons are interpreted as approximate short-to-mid-term anticipation horizons rather than exact wall-clock durations. Since durations are not explicitly modeled in the benchmark, all evaluation is reported in future steps rather than seconds.

\paragraph{\textbf{Scoring protocol}}
All headline metrics are computed at the cut position of each record. A benchmark property worth stating explicitly is the measured gap between this protocol and the all-steps protocol used by the predictability audit: scoring the same first-order model over all step positions rather than at the $r=0.50$ cut yields $+9.9$ points top-1 and $+6.0$ points top-5, so numbers across the two protocols must not be compared directly.

\subsection{Encoder Checkpoint}
\label{ssec:encoder_setup}

The recognition encoder is the graph NSGT of the companion work \cite{soleimani:hal-05616609}, operating on the typed episode graph over all four modalities. The checkpoint is fixed in advance to the project-wide default seed (seed 42, FOL condition \texttt{all}) rather than selected by score, avoiding best-seed selection bias; its full-observation recognition HLI top-1 on the test split is $69.7\%$, reproduced exactly (delta $0.000000$) by a reproducibility check before every anticipation run. The encoder is frozen throughout; its role as a reference point rather than a ceiling is discussed in Section~\ref{sec:results_j2}.

\subsection{Baselines}
\label{ssec:baselines}

Six comparison systems are evaluated, all consuming identical records and scored by one shared harness (itself gated by a fixture that reproduces stored baseline numbers to within $0.5$ points).

\paragraph{\textbf{B0, marginal frequency}} Predicts the globally most frequent actions, majority EOS, majority HLI, and the most frequent remaining sets. Establishes the floor.

\paragraph{\textbf{B1, first-order transition table}} A train-fitted transition table with greedy rollout from the last observed action; majority-style set and HLI heads.

\paragraph{\textbf{B2, bag-of-observed MLP}} A multi-task MLP over the unordered histogram of observed action labels, with heads for next actions, EOS, remaining sets, and HLI. This is the strongest compositional baseline.

\paragraph{\textbf{B3, sequential transformer}} A small autoregressive transformer over the observed label sequence, decoding next actions up to $K$ with EOS. This is the strongest sequential baseline.

\paragraph{\textbf{OntoPrior, symbolic-only}} A no-learning control: each observed action votes for every HLI whose ontology subtree contains it; next actions are the train-frequent unobserved actions inside the voted HLI subtree. By construction its predictions satisfy E1 and E2 exactly.

\paragraph{\textbf{LLM-LoRA}} The optional language-model decoder of Section~\ref{ssec:planning_decoder}, reported as a single ablation row.

B2, B3, and all HPD variants are trained with $3$ seeds and reported as mean with standard deviation; B0, B1, and OntoPrior are deterministic.

\subsection{Metrics}
\label{ssec:metrics}

\paragraph{\textbf{Next-action accuracy}} Step-$k$ top-1 and top-5 accuracy for $k \in \{1,2,3\}$ at the cut position, over the $81$-class effective vocabulary.

\paragraph{\textbf{Set-relaxed next-action}} Because the episode suffix is compositionally rather than ordinally determined, a set-relaxed variant credits a step-1 prediction that occurs anywhere in the unobserved suffix. Both exact-position and set-relaxed scores are reported.

\paragraph{\textbf{EOS}} F1 of the end-of-sequence flag.

\paragraph{\textbf{Remaining sets}} Jaccard similarity of the predicted versus ground-truth remaining-activity and remaining-LLI sets, complemented by set macro-mAP computed from the head scores.

\paragraph{\textbf{Goal inference}} HLI top-1 and macro-F1 at the cut, and HLI top-1 as a function of $r$.

\paragraph{\textbf{Logical coherence}} D1, D2 (report-only), E1, and E2 violation rates of the completed trajectory, read against the ground-truth floors of Section~\ref{ssec:benchmark}, and the episode-level satisfaction score over $\{$D1, E1, E2$\}$. D2 is reported for completeness but excluded from gating and headline claims: only $14$ of $50$ activities declare a canonical order, so the ground truth itself violates the canonical ordering on $22.4\%$ of test and $36.3\%$ of compositional trajectories, and a benchmark with denser ordering annotation would be needed to make within-activity ordering a discriminative target.

\section{Results and Analysis}
\label{sec:results_j2}

\subsection{Goal Inference under Partial Observation}
\label{ssec:results_goal}

\begin{figure}[!t]
\centering
\safeincludegraphics[width=\columnwidth]{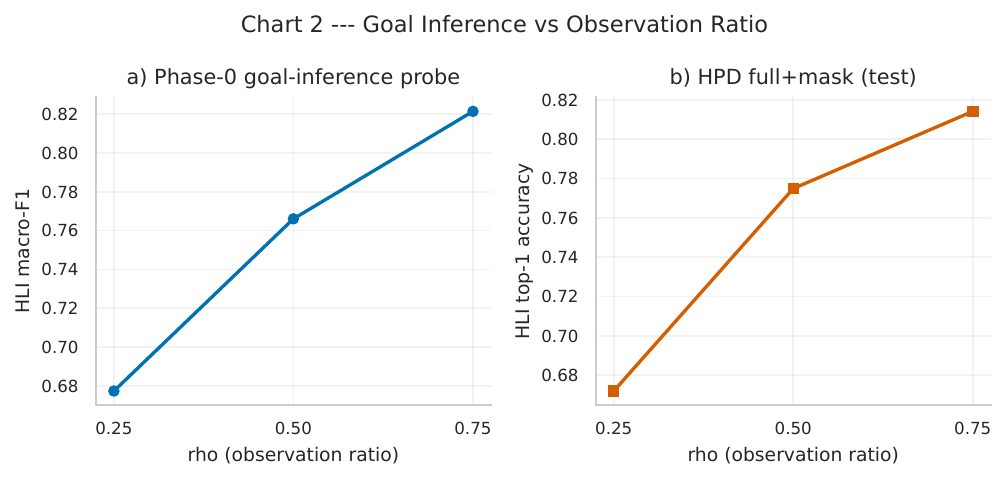}
\caption{Goal inference as a function of the observation ratio. The learned HPD tracks the label-only probe at short prefixes and exceeds it once half of the episode is observed.}
\label{fig:goal_vs_r}
\end{figure}

Table~\ref{tab:sensitivity_r} and Figure~\ref{fig:goal_vs_r} report the behavior of the gated HPD (fol\_mode full with constrained decoding) as the observation ratio varies over $r \in \{0.25, 0.50, 0.75\}$ on the test split, alongside the strongest sequential baseline and the label-only logistic probe of the predictability audit. Goal inference improves monotonically with observation: HLI top-1 rises from $67.2\%$ at a quarter of the episode to $81.4\%$ at three quarters. At $r = 0.25$ the learned system sits within half a point of the label-only probe, indicating that almost all goal evidence in a short prefix is carried by the identity of the observed actions; by $r = 0.50$ the HPD is clearly above the probe trajectory, indicating that the multimodal features and the decoder contribute goal evidence beyond the bare label sequence. All values are far above the $12.5\%$ chance level of the eight balanced HLI classes.

Step-level accuracy behaves differently. Step-1 top-5 increases monotonically ($79.7$, $81.5$, $89.3$), but step-1 top-1 dips at the mid-episode cut ($44.0$, $38.6$, $52.1$). This is a mechanical consequence of where the cut lands in an episode. At small $r$ the next action often still belongs to the activity already in progress in the prefix, so it is highly predictable from immediate context. At large $r$ the suffix is short and tends to consist of the frequent episode-closing actions. The mid-episode cut is the hardest case, because it sits simultaneously far from the opening context and from the closing pattern. This is the same effect measured as the protocol gap in the baseline phase, where scoring at all step positions rather than at the $r=0.50$ cut raised top-1 by $9.9$ points and top-5 by $6.0$ points; here that gap is resolved along the observation-ratio axis. The dip is therefore a property of cut position, not instability of the model.

A caveat applies to the full-observation reference. The frozen encoder reaches $69.7\%$ recognition HLI top-1 on full test episodes, yet the HPD reaches $77.5\%$ at $r = 0.50$. The recognition figure is a reference point, not a ceiling: by the protocol-parity decision of the study design, the anticipation task supplies the observed action labels as input, a signal the recognition task never receives. The label-only probe already reaches $76.6$ macro-F1 at $r = 0.50$ from those labels alone, so exceeding the encoder's recognition score reflects the difference in available inputs, not a contradiction.

\begin{table}[!t]
\caption{Sensitivity to the observation ratio $r$ on the test split (\%). Probe: deterministic label-only logistic probe of the predictability audit (HLI macro-F1). B3: mean over 3 seeds, computed from the per-seed sensitivity curves of the baseline report. HPD: fol\_mode full with constrained decoding, mean $\pm$ std over 3 seeds. Set-head masking affects only the set outputs and is irrelevant at this granularity.}
\label{tab:sensitivity_r}
\centering
\begin{tabular}{llccc}
\toprule
Model & Metric & $r{=}0.25$ & $r{=}0.50$ & $r{=}0.75$ \\
\midrule
Probe & HLI macro-F1 & 67.7 & 76.6 & 82.1 \\
\midrule
B3 seq-T & step-1 top-5 & 68.5 & 79.8 & 86.6 \\
B3 seq-T & HLI top-1    & 68.1 & 75.9 & 78.1 \\
\midrule
HPD full+mask & step-1 top-1 & 44.0$\pm$0.2 & 38.6$\pm$0.1 & 52.1$\pm$0.6 \\
HPD full+mask & step-1 top-5 & 79.7$\pm$0.6 & \textbf{81.5}$\pm$0.4 & \textbf{89.3}$\pm$0.5 \\
HPD full+mask & HLI top-1    & 67.2$\pm$1.1 & \textbf{77.5}$\pm$0.5 & \textbf{81.4}$\pm$0.9 \\
\bottomrule
\end{tabular}
\end{table}

\subsection{Next-Action Anticipation and Remaining Sets}
\label{ssec:results_antic}

\begin{figure}[!t]
\centering
\safeincludegraphics[width=\columnwidth]{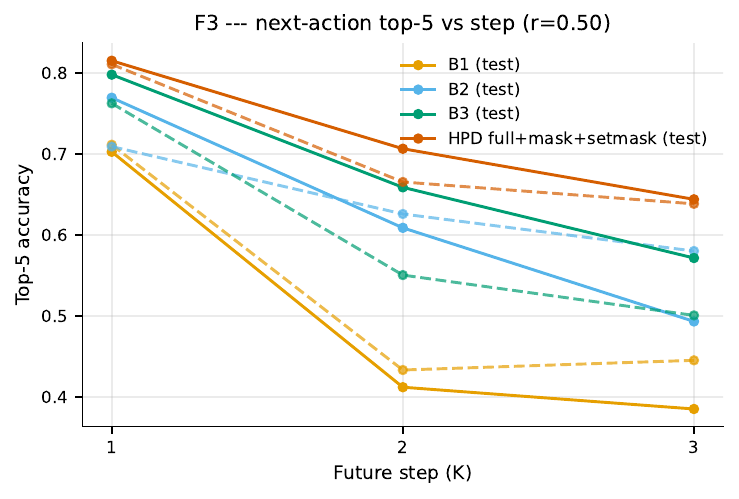}
\caption{Next-action top-5 versus future step at $r=0.50$. The margin of the gated HPD over the baselines widens with the horizon; dashed lines show the compositional split.}
\label{fig:f3_horizon}
\end{figure}

Table~\ref{tab:main_results} reports the primary results at $r = 0.50$. On the test split, the gated HPD reaches $81.5\%$ step-1 top-5, an improvement of $+1.7$ points over the strongest sequential baseline B3 ($79.8\%$) and $+1.1$ points over its own no-FOL ablation ($80.4\%$). The margin widens with the horizon: at step 3 the HPD leads B3 by $+7.3$ points ($64.4\%$ versus $57.1\%$, Table~\ref{tab:multistep_sat} and Figure~\ref{fig:f3_horizon}), indicating that the benefit of goal conditioning and constrained decoding compounds as the rollout departs from observed evidence.

\begin{figure}[!t]
\centering
\safeincludegraphics[width=\columnwidth]{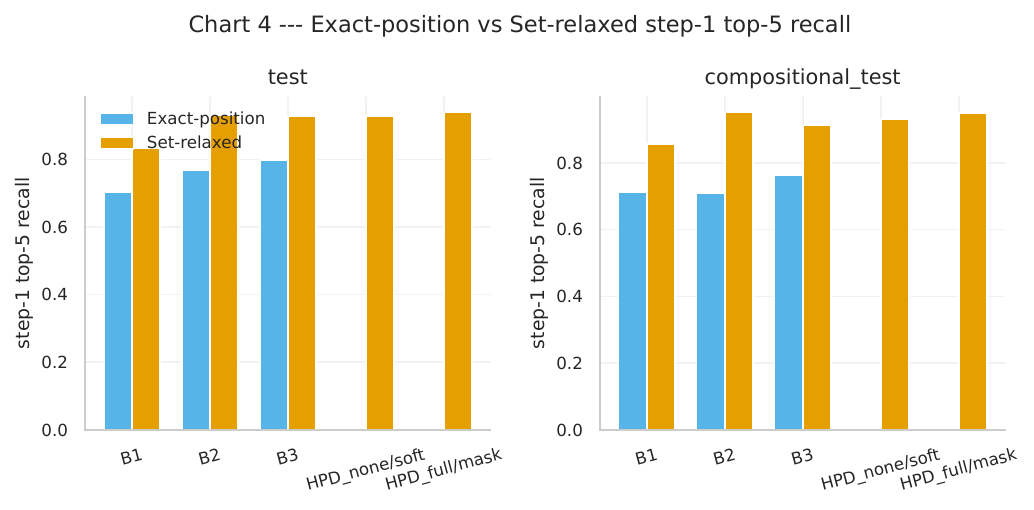}
\caption{Exact-position versus set-relaxed step-1 top-5. The large gap quantifies the compositional character of the remaining signal: future content is predicted far more reliably than its exact position.}
\label{fig:exact_vs_relaxed}
\end{figure}

The set-relaxed scores quantify the compositional nature of the remaining signal: crediting a step-1 prediction that occurs anywhere in the unobserved suffix lifts the HPD from $81.5$ to $93.8$ on test and from $81.1$ to $95.0$ on the compositional split (Figure~\ref{fig:exact_vs_relaxed}). The system therefore identifies future content considerably more reliably than its exact position, which is the expected behavior on data whose episode labels are order-insensitive while adjacent steps carry first-order structure.

Two honest observations from the same table. First, the ontology prior underperforms every learned model on next-action accuracy ($56.5\%$ top-5) despite satisfying E1 and E2 by construction: symbolic reachability restricts the candidate set but cannot rank within it, so learned generation contributes genuine predictive value beyond the ontology. Second, the histogram baseline B2 remains the best model for EOS and remaining-set prediction (EOS F1 $88.0$ versus $57.6$; activity Jaccard $40.1$ versus $22.3$), while being clearly weaker on sequential prediction and goal inference. The complementary-strengths pattern that motivated the HPD is thus only partially resolved: the decoder wins where sequencing and goal structure matter, and the unordered histogram wins where the target itself is a set. This limitation is taken up in Section~\ref{sec:discussion_j2}.

It is important to interpret these results through the compositional nature of the task. Exact-position top-1 accuracy remains moderate, because several future actions can be valid continuations of the same partially observed goal. For this reason, step-1 top-5 and set-relaxed top-5 are treated as the main next-action indicators, while exact top-1 is reported as a stricter diagnostic metric.

\paragraph{Comparison with egocentric anticipation benchmarks}
Absolute numbers on this benchmark are substantially higher than on natural egocentric anticipation datasets, where representative systems reach $16.5$ \cite{girdhar2021anticipative}, $23.8$ \cite{roy2024interaction}, and $39.7$ \cite{assran2025v} mean top-5 recall on EPIC-KITCHENS-100. These figures must not be compared with Table~\ref{tab:main_results}. The benchmark used here is synthetic with strong compositional and transition structure, its effective vocabulary is $81$ classes against $3{,}568$ EPIC verb-noun classes, and the metrics differ (accuracy here versus class-mean recall there). The contribution of this article is the controlled attribution of hierarchical, symbolic, and feature components, not a state-of-the-art claim on natural video.

\begin{table*}[t]
\caption{Anticipation at $r=0.50$ over the 81-class effective vocabulary (values in \%). Top-1/Top-5: exact-position step-1 next action. Jacc: remaining-set Jaccard. D1/E1/E2: violation rates of the completed trajectory; ground-truth floors are D1 $=6.4$ test and $0.7$ compositional, E1 $=$ E2 $=0$ on both splits; D2 is report-only (Section IV) and shown in Fig.~\ref{fig:f4_viol}. B2, B3, and HPD rows are means over 3 seeds; B0, B1, and OntoPrior are deterministic; LLM-LoRA is a single seed. { \textbf{Per-seed standard deviations for all learned configurations are reported in Table~\ref{tab:appendix_std}.}} $^{a}$The train-majority HLI class is absent from the compositional split by construction of the pattern-based holdout, so majority-vote HLI scores zero; this is a split property, not a defect. $^{b}$LLM-LoRA generates a single candidate, so top-5 equals top-1, and it never emits the END token, so EOS F1 is zero.}
\label{tab:main_results}
\centering
\adjustbox{max width=\textwidth}{%
\begin{tabular}{llccccccccc}
\toprule
Model & Split & Top-1 & Top-5 & EOS F1 & HLI top-1 & Act Jacc & LLI Jacc & D1 & E1 & E2 \\
\midrule
B0 marginal    & test & 4.3  & 33.8 & 75.2 & 11.4 & 19.8 & 7.8  & 7.1 & 0.0 & 0.0 \\
B1 transition  & test & 26.3 & 70.3 & 75.2 & 11.4 & 19.8 & 7.8  & 0.7 & 26.8 & 0.0 \\
B2 bag-MLP     & test & 34.9 & 76.9 & 88.0 & 77.3 & 40.1 & 40.2 & 7.7 & 2.5 & 0.2 \\
B3 seq-T       & test & 38.6 & 79.8 & 64.0 & 75.9 & 30.4 & 32.0 & 1.5 & 5.5 & 0.1 \\
OntoPrior      & test & 11.2 & 56.5 & 75.2 & 67.4 & 24.4 & 18.4 & 6.0 & 0.0 & 0.0 \\
LLM-LoRA$^{b}$ & test & 16.8 & 16.8 & 0.0  & 30.1 & 22.7 & 22.1 & 2.7 & 3.2 & 2.1 \\
HPD none (soft) & test & 38.1 & 80.4 & 56.7 & 78.2 & 22.6 & 32.1 & 1.6 & 5.0 & 0.8 \\
HPD full+mask+setmask & test & 38.6 & \textbf{81.5} & 57.6 & 77.5 & 22.3 & 31.0 & 1.1 & \textbf{0.0} & \textbf{0.0} \\
\midrule
B1 transition  & comp. & 27.9 & 71.1 & 69.0 & 0.0$^{a}$ & 24.4 & 11.6 & 0.0 & 16.1 & 0.0 \\
B2 bag-MLP     & comp. & 25.4 & 70.9 & 85.9 & 53.1 & 25.1 & 14.0 & 2.1 & 1.6 & 0.0 \\
B3 seq-T       & comp. & 35.1 & 76.2 & 60.7 & 52.3 & 14.1 & 5.8  & 0.9 & 10.8 & 0.0 \\
OntoPrior      & comp. & 13.9 & 58.7 & 69.0 & 39.4 & 20.0 & 14.1 & 0.9 & 0.0 & 0.0 \\
LLM-LoRA$^{b}$ & comp. & 18.1 & 18.1 & 0.0  & 25.3 & 22.9 & 17.8 & 2.4 & 3.7 & 2.1 \\
HPD none (soft) & comp. & 34.7 & 78.6 & 56.9 & \textbf{55.9} & 3.4 & 10.5 & 0.1 & 7.7 & 0.4 \\
HPD full+mask+setmask & comp. & 35.1 & \textbf{81.1} & 58.1 & 53.9 & 4.2 & 10.6 & 0.0 & \textbf{0.0} & \textbf{0.0} \\
\bottomrule
\end{tabular}}
\end{table*}

\subsection{Attribution: Architecture, Features, and Logic}
\label{ssec:results_attribution}

\begin{figure}[!t]
\centering
\safeincludegraphics[width=\columnwidth]{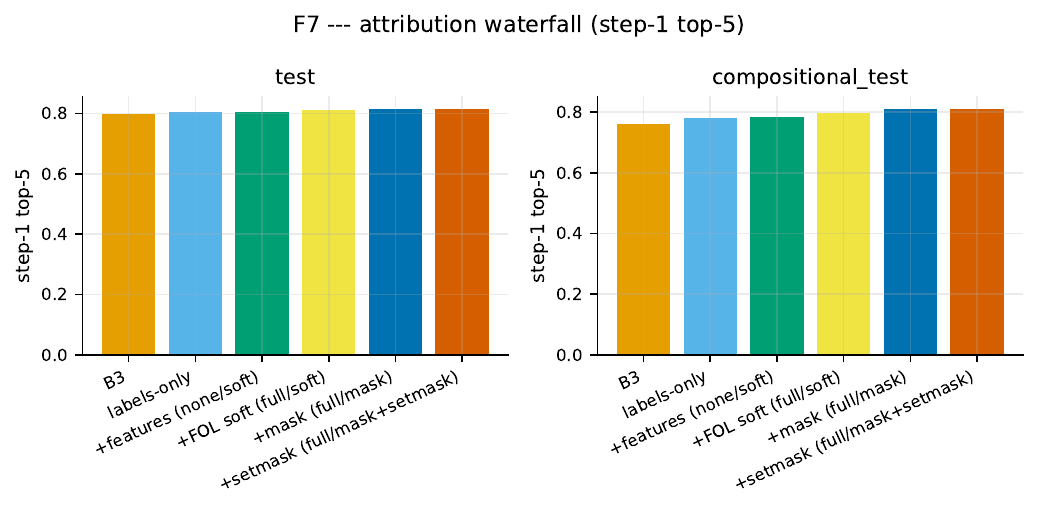}
\caption{Attribution ladder for step-1 top-5: B3, labels-only HPD, HPD with multimodal features, and the full gated system, on both splits.}
\label{fig:f7_ladder}
\end{figure}

The improvement over B3 decomposes into three separable contributions, measured by an input-ablation ladder evaluated under identical protocol (Figure~\ref{fig:f7_ladder}). Starting from B3 ($79.8$ test, $76.2$ compositional step-1 top-5), replacing the baseline with the HPD architecture restricted to the same label-only input yields $80.5$ and $78.2$: pure architecture contributes modestly on test ($+0.7$) and more visibly on the compositional split ($+2.0$), where cross-attention over the prefix generalises better than the baseline's flat encoding. 
Adding the frozen multimodal encoder states leaves next-action accuracy essentially unchanged ($80.4$ and $78.6$) but raises HLI top-1 by $+2.7$ points on both splits and improves remaining-set prediction. A mechanistic reading is available for this asymmetry: the benchmark synthesizes episode adjacency at the label level, so the observed action labels already carry the local dynamics that next-action ranking depends on; the multimodal encoder states contribute orthogonal appearance and pose evidence, which is informative for the goal (HLI $+2.7$ points) but not for the next label itself. Finally, activating the extended FOL loss with constrained decoding yields the full system at $81.5$ and $81.1$. The set macro-mAP tells the same story from the set side: the full HPD reaches $28.2$ activity and $37.0$ LLI mAP on test against $38.8$ and $45.8$ for the histogram baseline B2, confirming that the histogram remains the stronger set predictor.

\subsection{Logic Coherence of Anticipated Trajectories}
\label{ssec:results_fol}

\begin{figure*}[!t]
\centering
\safeincludegraphics[width=0.48\textwidth]{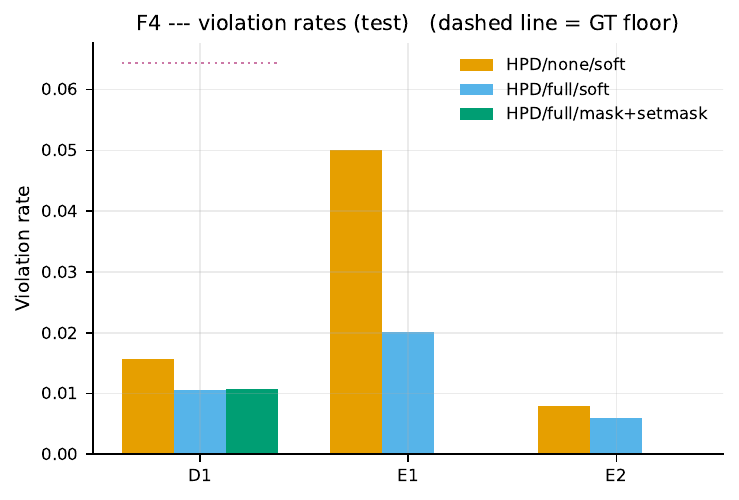}\hfill
\safeincludegraphics[width=0.48\textwidth]{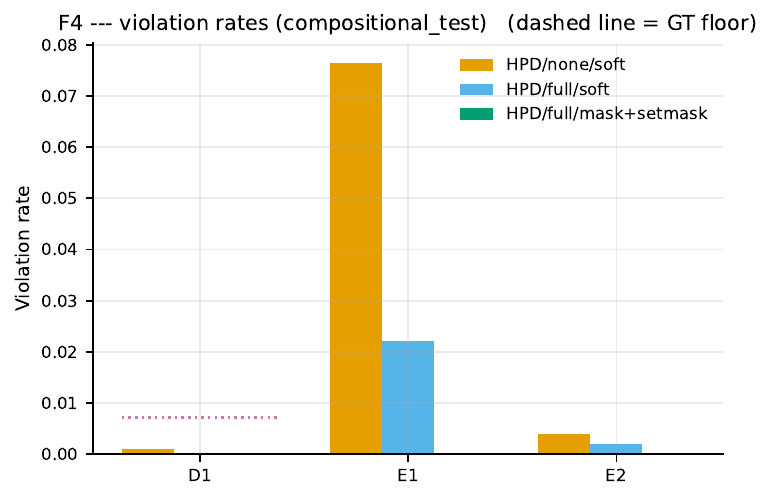}
\caption{Violation rates of anticipated trajectories at $r=0.50$ on the test split (left) and the compositional split (right), for the no-FOL HPD, the soft-trained system, and the final gated configuration. Dotted lines mark the intrinsic ground-truth floors.}
\label{fig:f4_viol}
\end{figure*}

Figure~\ref{fig:f4_viol} and the E1/E2 columns of Table~\ref{tab:main_results} report the coherence results, which are stated in two layers to keep attribution honest.

\paragraph{Soft training alone}
With no logic term, the HPD violates HLI-reachability (E1) on $5.0\%$ of test trajectories and $7.7\%$ of compositional ones. Training with the soft Type-D and Type-E relaxations, with unconstrained decoding, reduces E1 to $2.0\%$ and $2.2\%$, relative reductions of $59.8\%$ and $71.1\%$. The soft loss alone therefore moves the model most of the way, and the reduction is larger exactly where generalization is hardest.

\paragraph{Constrained decoding}
Adding the inference-time reachability mask eliminates E1 entirely ($0.0\%$ on both splits) while simultaneously improving accuracy: step-1 top-5 rises by $+1.1$ points and HLI top-1 moves by less than $0.7$ points relative to the unconstrained no-FOL system. Coherence is therefore obtained without accuracy degradation. The mask initially left the set heads exposed: E2 of the masked system was $0.6\%$ on test, missing the $0.5\%$ secondary ceiling by $0.09$ points. Extending the same reachability filter to the set heads (setmask) drives E2 to exactly $0.0\%$ on both splits at no measurable accuracy change, and this configuration, full+mask+setmask, is the system reported everywhere in this article. Its D1 rate ($1.1\%$ test) is below that of the no-FOL ablation ($1.6\%$) and far below the ground-truth floor of $6.4\%$; predicted trajectories concentrate on well-supported transitions. D2 is not discriminative on this benchmark, with an intrinsic floor of $22.4\%$ test and $36.3\%$ compositional, and is reported in Figure~\ref{fig:f4_viol} for completeness only.

At the episode level, $96.83\%$ of test trajectories of the final system carry zero violations over $\{$D1, E1, E2$\}$, against $88.09\%$ for the best baseline and a ground-truth floor of $73.92\%$ (Table~\ref{tab:multistep_sat}). Each satisfaction figure must be read against the floor of its own split: on the compositional split the floor itself is $93.79\%$, and the final system reaches $99.96\%$.

\begin{table}[t]
\caption{Horizon decay (top-5, exact position), set-relaxed step-1 top-5, and episode-level satisfaction over $\{$D1, E1, E2$\}$ at $r=0.50$. Mean $\pm$ std over 3 seeds; satisfaction reported to two decimals because the HPD reaches 99.96\% on the compositional split.}
\label{tab:multistep_sat}
\centering
\setlength{\tabcolsep}{3.0pt}
\begin{tabular}{llccccc}
\toprule
Model & Split & Step 1 & Step 2 & Step 3 & Set-rel. & Satisf. \\
\midrule
B3 seq-T & test & 79.8$\pm$0.3 & 65.9$\pm$0.7 & 57.1$\pm$1.1 & 92.7$\pm$0.8 & 88.09$\pm$0.73 \\
HPD f+m+s & test & 81.5$\pm$0.4 & 70.6$\pm$0.3 & 64.4$\pm$0.8 & 93.8$\pm$0.3 & 96.83$\pm$0.14 \\
\midrule
B3 seq-T & comp. & 76.2$\pm$2.0 & 55.0$\pm$1.3 & 50.1$\pm$1.3 & 91.2$\pm$2.4 & 83.76$\pm$1.39 \\
HPD f+m+s & comp. & 81.1$\pm$1.4 & 66.5$\pm$0.9 & 63.8$\pm$1.6 & 95.0$\pm$0.1 & 99.96$\pm$0.04 \\
\midrule
GT floor & test & \multicolumn{4}{c}{} & 73.92 \\
GT floor & comp. & \multicolumn{4}{c}{} & 93.79 \\
\bottomrule
\end{tabular}
\end{table}

\subsection{Compositional Generalization}
\label{ssec:results_comp}

\begin{figure}[!t]
\centering
\safeincludegraphics[width=\columnwidth]{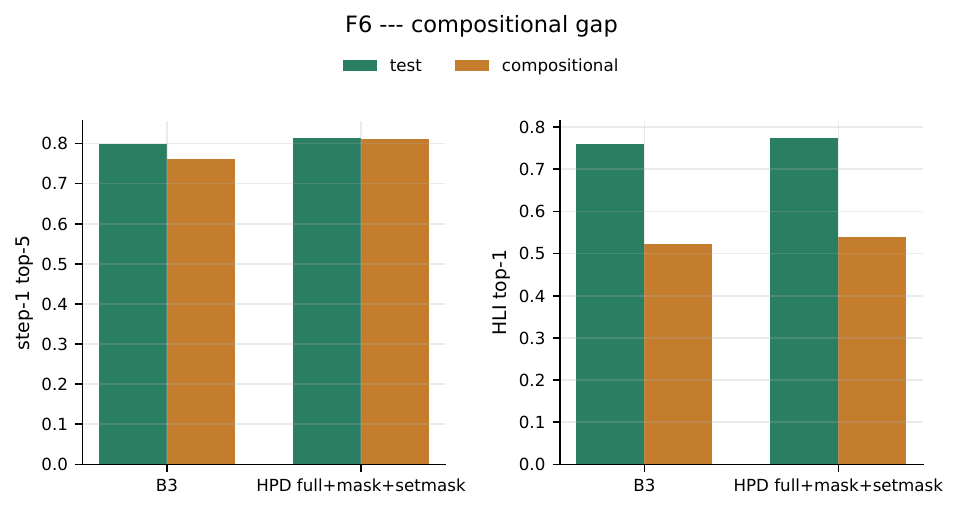}
\caption{Test versus compositional split for step-1 top-5 and HLI top-1, B3 and the final HPD. Sequential accuracy transfers; goal inference carries a gap of roughly 24 points.}
\label{fig:f6_comp}
\end{figure}

The compositional split withholds one parent association per multi-parent LLI, so goal inference must transfer a concept to a high-level context never seen in training. Figure~\ref{fig:f6_comp} summarises the outcome. Goal inference inherits the architecture-independent gap documented for recognition on this benchmark: HLI top-1 drops from $77.5$ to $53.9$ for the final system and from $77.3$ to $53.1$ for B2, a gap of roughly $24$ points in both cases, and this gap remains the dominant open problem. Sequential prediction, by contrast, transfers almost unchanged: step-1 top-5 moves from $81.5$ to $81.1$, and the margin over B3 grows from $+1.7$ to $+4.9$ points, because local transition structure is shared across splits while goal composition is not. The set heads collapse on the held-out compositions (activity Jaccard $22.3$ to $4.2$), a direct consequence of predicting sets for goal contexts whose composition was never observed. Logical coherence transfers best of all: E1 and E2 remain exactly zero, and episode satisfaction reaches $99.96\%$ against the split's own floor of $93.79\%$. Coherence guarantees provided by the constrained decoder are structural and therefore survive distribution shift that accuracy does not.

\subsection{Language-Model Decoder Row}
\label{ssec:results_llm}

The optional language-model decoder (Qwen2.5-1.5B-Instruct, 4-bit base, LoRA rank 16 on all attention and MLP projections, $18.5$M trainable parameters, single seed) is trained to emit the full structured target as JSON from a textual rendering of the observed action names. Format compliance is high: the parse rate is $1.00$ on both splits and the inferred HLI appears in the generated explanation in every case. Task accuracy is not competitive: $16.8\%$ step-1 top-1 on test against $38.6\%$ for the HPD, and $30.1\%$ HLI top-1 against $77.5\%$, with a violation profile (E1 $3.2\%$, E2 $2.1\%$) that no component of the pipeline constrains. Two properties of this row require explicit qualification. First, the model generates a single candidate sequence, so its top-5 equals its top-1 and the top-5 column understates nothing. Second, the reported BLEU-4 of $78.9$ test and $74.6$ compositional is computed against a synthetic explanation template that also served as the fine-tuning target; it measures template compliance, not explanation quality, and is reported only to document that the language interface is functional. On this benchmark, whose vocabulary is a closed set of numeric label ids rendered as names, prompt-based generation offers no advantage over the structured decoder; whether language priors help on naturally textual vocabularies is left outside the scope of these experiments.

{ \subsection{Robustness to Predicted Observed Labels}
\label{ssec:results_robustness}

The main configuration reports numbers under an oracle-label protocol, matching the label-oracle baselines. This subsection evaluates what happens when the oracle assumption is relaxed. All results below use the same trained final configuration (full+mask+setmask) reported in Table~\ref{tab:main_results}; the values there correspond to the oracle row.

\paragraph{\textbf{Encoder-predicted labels (condA)}}
Replacing every observed action label with the encoder-argmax prediction gives a measured label accuracy of $82.13\%$ over the $24{,}764$ observed positions across test and compositional test at $r=0.50$. Under these predicted labels, the clean-trained final system retains $71.4\%$ step-1 top-5 (down from $81.5\%$) and $67.3\%$ HLI top-1 (down from $77.5\%$) on the test split, with step-3 top-5 nearly unaffected ($61.0$ versus $64.4$) and the remaining-set heads mildly affected (activity Jaccard $-1.5$, LLI Jaccard $-4.2$). On the compositional split, HLI top-1 drops from $53.9\%$ to $43.2\%$ while set-Jaccard values move by less than $1.5$ points. Critically, E1 and E2 remain exactly $0\%$ under every condition tested: the reachability guarantee is structural, not a function of input quality.

\paragraph{\textbf{Synthetic label-noise sweep (condB)}}
Perturbing each observed label independently with probability $p \in \{0.10, 0.20, 0.30\}$ produces monotonic, well-behaved degradation. Step-1 top-5 on test moves from $78.5$ to $74.3$ to $71.6$; HLI top-1 from $75.4$ to $73.5$ to $71.2$; step-3 top-5 stays within one point of the oracle across the range. Two contrasts are informative. First, condA has a measured label-error rate of $17.87\%$, closest to the $p{=}0.20$ point of the sweep; at that matched rate, iid noise degrades HLI top-1 on test by $-4.0$ points, while the encoder's real correlated errors degrade it by $-10.2$ points. The $6$-point gap under a matched error rate says that the composition of the errors, not their frequency, is what hurts goal inference: encoder confusions cluster around visually similar classes and can concentrate several within a single episode, whereas iid corruption spreads the same total error uniformly across positions. Correlated encoder errors, which tend to cluster around visually confusable classes and can concentrate within an episode, hurt goal inference substantially more than iid noise of the same rate. Second, E1 and E2 remain $0\%$ throughout the sweep, again confirming that the coherence guarantee is independent of input quality.

\paragraph{\textbf{Noise-aware training (deployment variant)}}
A second variant of the final configuration is trained with a $50\%$ probability of replacing each observed label with the encoder prediction during training, keeping every other choice identical. On test, this variant improves the fully end-to-end (encoder-predicted-label) condition by $+3.8$ points step-1 top-5 and $+1.6$ points HLI top-1; on the compositional split by $+4.0$ and $+2.8$ points respectively. The oracle-label numbers of the same trained model are within seed noise of the clean-trained configuration ($-0.4$ points step-1 top-5, $-1.4$ points HLI top-1 on test), so noise-aware training essentially closes the deployment gap at a negligible oracle cost. E1 and E2 remain exactly $0\%$ in every condition, oracle or predicted, confirming again that the reachability guarantee holds regardless of the training protocol. The main system of this article stays the clean-trained configuration reported in Table~\ref{tab:main_results}; the noise-aware variant is reported here as a deployment-oriented robustness row (Table~\ref{tab:robustness}) and is the recommended configuration when only encoder-predicted labels are available at test time.

\begin{table}[!t]
\caption{Robustness to predicted observed labels at $r=0.50$ (final full+mask+setmask configuration; mean $\pm$ std over 3 seeds). Oracle = observed labels are the ground-truth NTU indices (main-paper protocol). Cond A = observed labels are the frozen recognition encoder's argmax predictions (measured accuracy $82.13\%$). Cond B $p$ = each observed label is independently corrupted with probability $p$. \emph{Clean-trained} = the model of Table~\ref{tab:main_results}. \emph{Noise-aware} = the same architecture trained with a $50\%$ probability of encoder-predicted labels. E1 and E2 are exactly $0.00 \pm 0.00$ in every cell and are omitted from the table for brevity. The EOS F1 and LLI Jaccard columns are also omitted here for readability; the full column set for every row is available in the appendix (Table~\ref{tab:appendix_robustness_full}).}
\label{tab:robustness}
\centering
\setlength{\tabcolsep}{2.4pt}
\adjustbox{max width=\columnwidth}{%
\begin{tabular}{lllcccc}
\toprule
Training & Condition & Split & step-1 top-5 & step-3 top-5 & HLI top-1 & act.\ Jacc \\
\midrule
Clean       & oracle       & test  & 81.5$\pm$0.4 & 64.4$\pm$0.8 & 77.5$\pm$0.5 & 22.3$\pm$1.6 \\
Clean       & condA        & test  & 71.4$\pm$0.5 & 61.0$\pm$1.1 & 67.3$\pm$0.1 & 20.7$\pm$1.4 \\
Clean       & condB $p{=}0.10$ & test & 78.5$\pm$0.7 & 63.4$\pm$0.6 & 75.4$\pm$0.2 & 21.0$\pm$1.6 \\
Clean       & condB $p{=}0.20$ & test & 74.3$\pm$0.6 & 61.9$\pm$0.8 & 73.5$\pm$0.2 & 19.5$\pm$1.4 \\
Clean       & condB $p{=}0.30$ & test & 71.6$\pm$0.6 & 60.7$\pm$0.3 & 71.2$\pm$0.2 & 18.4$\pm$1.3 \\
Noise-aware & oracle       & test  & 81.1$\pm$0.6 & 63.8$\pm$1.5 & 76.1$\pm$0.4 & 22.2$\pm$1.3 \\
Noise-aware & condA        & test  & 75.2$\pm$1.1 & 60.7$\pm$1.4 & 68.9$\pm$0.4 & 20.8$\pm$1.0 \\
\midrule
Clean       & oracle       & comp. & 81.1$\pm$1.4 & 63.8$\pm$1.6 & 53.9$\pm$1.0 & 4.2$\pm$1.1 \\
Clean       & condA        & comp. & 72.2$\pm$1.6 & 61.3$\pm$1.7 & 43.2$\pm$0.7 & 4.1$\pm$1.0 \\
Clean       & condB $p{=}0.10$ & comp. & 76.8$\pm$1.4 & 62.9$\pm$1.2 & 52.1$\pm$1.0 & 3.9$\pm$0.9 \\
Clean       & condB $p{=}0.20$ & comp. & 73.0$\pm$1.8 & 62.5$\pm$1.2 & 49.3$\pm$0.8 & 3.5$\pm$0.8 \\
Clean       & condB $p{=}0.30$ & comp. & 69.1$\pm$1.4 & 61.2$\pm$1.6 & 47.6$\pm$0.8 & 3.2$\pm$0.8 \\
Noise-aware & oracle       & comp. & 81.6$\pm$0.3 & 62.9$\pm$1.4 & 53.3$\pm$1.2 & 5.0$\pm$2.5 \\
Noise-aware & condA        & comp. & 76.3$\pm$1.0 & 62.0$\pm$1.8 & 46.0$\pm$1.2 & 4.6$\pm$2.1 \\
\bottomrule
\end{tabular}}
\end{table}}

\begin{figure}[!t]
\centering
\resizebox{\columnwidth}{!}{%
\begin{tikzpicture}[
  actbox/.style={draw, rounded corners=2pt, minimum height=5.5mm,
    inner sep=2pt, font=\scriptsize, text centered},
  obsbox/.style={actbox, fill=obsgray!20, draw=obsgray},
  gtbox/.style={actbox, fill=softgreen!18, draw=softgreen!80!black},
  b3box/.style={actbox, fill=hardred!15, draw=hardred},
  hpbox/.style={actbox, fill=encblue!12, draw=encblue},
  arr/.style={-{Stealth[length=2.5pt]}, thick, black!45},
  rowlbl/.style={font=\footnotesize\bfseries, anchor=east},
  hlilbl/.style={font=\scriptsize, anchor=west},
]
\def\rO{0}
\def\rG{-1.3}
\def\rB{-2.7}
\def\rH{-4.0}
\node[rowlbl] at (-0.3,\rO) {Observed};
\node[obsbox] (o1) at (1.5,\rO) {open bottle};
\node[obsbox] (o2) at (3.3,\rO) {fan self};
\node[obsbox] (o3) at (5.0,\rO) {phone call};
\node[obsbox] (o4) at (7.1,\rO) {play w/ phone};
\node[obsbox] (o5) at (8.8,\rO) {yawn};
\node[obsbox] (o6) at (10.5,\rO) {stretch};
\foreach \a/\b in {o1/o2,o2/o3,o3/o4,o4/o5,o5/o6}
  \draw[arr] (\a)--(\b);
\draw[dashed, black!30] (-0.6,-0.65) -- (12.8,-0.65);
\node[rowlbl] at (-0.3,\rG) {GT};
\node[gtbox] (g1) at (1.5,\rG) {stand up};
\node[gtbox] (g2) at (3.4,\rG) {open a box};
\node[gtbox] (g3) at (5.6,\rG) {put on jacket};
\draw[arr] (g1)--(g2); \draw[arr] (g2)--(g3);
\node[hlilbl,softgreen!60!black] at (7.4,\rG)
  {\textsc{hli}: MorningRoutine\;\checkmark};
\node[rowlbl] at (-0.3,\rB) {B3};
\node[b3box] (b1) at (1.7,\rB) {put on jacket};
\node[b3box] (b2) at (3.9,\rB) {put on a shoe};
\node[b3box] (b3) at (6.3,\rB) {put on a hat/cap};
\draw[arr] (b1)--(b2); \draw[arr] (b2)--(b3);
\node[font=\tiny,hardred,anchor=south] at (b1.north) {\textbf{E1!}};
\node[font=\tiny,hardred,anchor=south] at (b2.north) {\textbf{E1!}};
\node[font=\tiny,hardred,anchor=south] at (b3.north) {\textbf{E1!}};
\node[hlilbl,hardred!80!black] at (8.2,\rB)
  {\textsc{hli}: OfficeWorkSession};
\node[font=\scriptsize,hardred!80!black,anchor=north west] at (8.2,\rB-0.15)
  {violates E1\,$\times\!3$};
\node[rowlbl] at (-0.3,\rH) {HPD};
\node[hpbox] (h1) at (1.7,\rH) {open bottle};
\node[hpbox] (h2) at (3.9,\rH) {cheers \& drink};
\node[hpbox] (h3) at (5.8,\rH) {sit down};
\draw[arr] (h1)--(h2); \draw[arr] (h2)--(h3);
\node[hlilbl,encblue!80!black] at (7.4,\rH)
  {\textsc{hli}: OfficeWorkSession\;\checkmark};
\node[font=\scriptsize,encblue!70!black,anchor=north west] at (7.4,\rH-0.15)
  {E1\,$=0$ (mask-consistent)};
\end{tikzpicture}%
}
\caption{Qualitative comparison on episode \texttt{ep\_0000331} (deterministically selected: the first test episode on which B3 violates HLI-reachability while the gated HPD remains consistent). B3 and HPD both mis-predict HLI as \emph{OfficeWorkSession} (GT: \emph{MorningRoutine}), yet B3 emits three dressing actions unreachable from its predicted HLI (three E1 violations), while HPD's reachability mask forces it to select office-compatible actions (E1\,$=0$).}
\label{fig:f8_qual}
\end{figure}
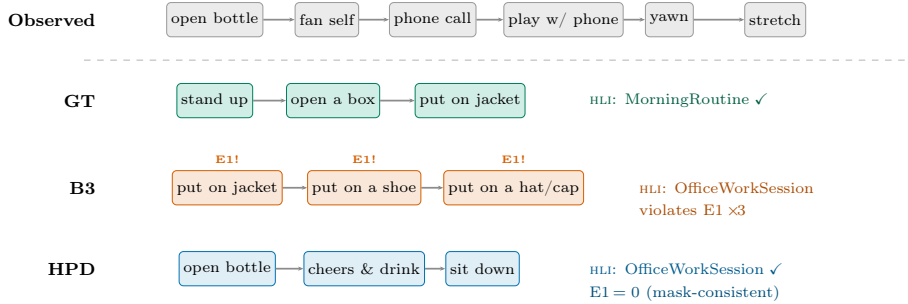

\section{Discussion and Limitations}
\label{sec:discussion_j2}

\subsection{Core Empirical Findings}

Three findings are supported by the experiments under the stated synthetic-benchmark protocol. First, goal inference from a partial episode is feasible and improves smoothly with observation, reaching $77.5\%$ HLI top-1 at half an episode and $81.4\%$ at three quarters, well above chance and above the frozen encoder's recognition reference once the input-parity difference is accounted for. Second, the soft/hard neuro-symbolic separation works as designed: the soft FOL loss removes $59.8$ to $71.1\%$ of reachability violations during training, and inference-time reachability masking closes the remaining gap to exactly zero while improving accuracy rather than trading it away. Third, structural coherence guarantees survive distribution shift that accuracy does not: on the compositional split, goal inference loses roughly $24$ points while reachability violations stay at zero and episode satisfaction stays above its own ground-truth floor.

\subsection{Baseline Evidence for the Neuro-Symbolic Architecture}

The clearest argument for combining neural generation with symbolic constraints comes from a three-way contrast that the baselines make explicit. The purely symbolic OntoPrior, which completes trajectories by ontology reachability alone, satisfies E1 and E2 by construction but is weak at prediction: its next-action top-5 is only $56.5\%$, far below every learned model, because reachability restricts the candidate set without ranking within it. The purely neural HPD without any logic, in contrast, is strong at prediction ($80.4\%$ top-5) but violates HLI-reachability on $5.0\%$ of trajectories, because nothing constrains its generation to the ontology. The combination of the two, soft logic during training plus hard masking at inference, is not a compromise between these but strictly better than either: it matches or exceeds the neural model's accuracy ($81.5\%$) while achieving perfect reachability consistency. The episode-level satisfaction figure sharpens the point: the combined system satisfies the joint constraint set on $96.8\%$ of test trajectories, exceeding not only the best baseline ($88.1\%$) but the ground-truth floor itself ($73.9\%$), because the constrained decoder cannot emit the unsupported transitions that even the synthesized ground-truth trajectories contain. Neither the symbolic nor the neural component reaches this alone.

\subsection{Internalization of Logical Constraints}
\label{ssec:internalized_logic}

It is necessary to determine the extent to which the soft FOL loss causes the decoder to internalize the ontology, as opposed to the inference-time mask simply projecting an unconstrained decoder onto the reachable set. This distinction is isolated through the FOL-mode ablation. When training is performed with the soft Type-D and Type-E terms alongside unconstrained decoding, HLI-reachability violations are reduced from $5.00\%$ to $2.01\%$ on the test split and from $7.65\%$ to $2.20\%$ on the compositional split. These figures represent relative reductions of $59.8$ and $71.1$ percent, which are achieved without relying on any decoding-time filter. Consequently, the model is independently guided toward logical coherence by the soft loss, and this effect is particularly pronounced in scenarios where generalization is more challenging. The remaining fraction of violations is eliminated by the inference-time reachability mask, driving the E1 metric to $0.00\%$ on both splits without any degradation in accuracy. Therefore, two primary conclusions are derived.

First, the network does internalize part of the ontology under the soft loss alone: without the inference-time mask, HLI-reachability violations are substantially reduced, from $5.00\%$ to $2.01\%$ on the test split and from $7.65\%$ to $2.20\%$ on the compositional split.
Second, the mask is necessary for a hard guarantee: the residual $2$-percent-scale violations that the soft loss does not remove are exactly what the mask eliminates. The combination is not redundant, and neither part is doing the other's job.

\subsection{Reconciling Order-Insensitivity with Sequential Predictability}

A tension that appears contradictory is in fact two design properties of the benchmark. Episode-level classification is order-insensitive: the order-destroying control shifts recognition scores only within seed noise, because an episode label is fixed by an unordered composition of parts. Yet next-step prediction carries a strong first-order signal: conditioning on the previous action lifts next-action top-5 by $38.7$ points over the marginal predictor and lowers the conditional entropy of the next action from $5.21$ to $3.51$ bits. Both hold at once because adjacent actions are generated by a transition model, while the goal that the episode realises is not a function of their order. The task design mirrors this exactly, pairing sequential next-action prediction, where local order is informative, with set-valued and episode-level targets, where it is not. The large gap between exact-position and set-relaxed next-action scores ($81.5$ versus $93.8$ top-5) is the quantitative footprint of the same property: the system frequently knows what is coming without knowing precisely when.

\subsection{Limitations}

The benchmark composes real multimodal features into synthesized episode structure with a released transition model and typed FOL rules, so the findings of this article are methodological and the absolute numbers do not transfer to naturally recorded long-horizon behavior; they are not directly comparable with anticipation results on egocentric video, whose vocabulary, metrics, and generative processes differ substantially. External transfer was audited and excluded: a $\psi$-align vocabulary audit against EPIC-KITCHENS-100 found strong correspondence for $6.7\%$ of classes and no usable correspondence for $76.2\%$, ruling out direct reuse of the encoder or the anticipation vocabulary. Among the alternatives, Ego4D lacks the required skeleton, depth, and infrared streams; CAD-120 offers hierarchy at only about $120$ videos; Breakfast, 50 Salads, and Toyota Smarthome fail on vocabulary or on missing goal labels.

The roughly $24$-point drop in HLI top-1 on the held-out compositional split is the dominant unresolved problem, and it is not a peculiarity of the proposed system. On the same benchmark, the recognition companion \cite{soleimani:hal-05616609} reports a full-observation HLI macro-F1 gap of $12.7$ to $17.2$ points across four recognition baselines, and under the partial-observation protocol used here the gap widens to about $24$ points at $r=0.50$ for every model, from the histogram baseline B2 ($77.3$ to $53.1$) to the final HPD ($77.5$ to $53.9$), while sequential next-action prediction remains nearly gap-free ($81.5$ to $81.1$ step-1 top-5); the two protocols use different metrics and observation regimes, so only the direction and magnitude of the gap are being compared. The full evidence is in Section~\ref{ssec:results_comp}.

Set-valued targets remain led by the bag-of-observed histogram baseline B2, which outperforms every learned decoder on EOS detection and remaining-set Jaccard while being clearly weaker on sequential prediction and goal inference (Table~\ref{tab:main_results}); a unified module that closes both fronts cleanly is not yet in hand. The recognition encoder is frozen after training on full episodes and is therefore not adapted to prefix-based inputs, which likely limits the low-observation regime; the noise-aware training row of Section~\ref{ssec:results_robustness} closes most of the gap under encoder-predicted labels without touching the encoder, but end-to-end adaptation of the encoder to partial-observation prefixes remains untested. Explicit relational generalization for the compositional goal gap, unified handling of set-valued and sequential targets, and encoder adaptation to prefixes are the three concrete directions this work leaves open.

\section{Conclusion}
\label{sec:conclusion_j2}

This article formulated hierarchical behavior anticipation as goal inference from a partially observed multimodal episode together with structured prediction of the remaining behavior. A compact Hierarchical Planning Decoder was attached to a frozen neuro-symbolic recognition encoder and trained with soft transition-coherence and hierarchical-continuity losses, while hard reachability masks enforced ontological validity at inference. On the four-level compositional benchmark, the final system improved next-action top-5 accuracy over the strongest sequential baseline, reached strong HLI inference at half-episode observation, and eliminated HLI-reachability violations without degrading accuracy.

The results support the value of combining neural generation with symbolic structure: symbolic reachability alone is coherent but weak, a purely neural decoder is accurate but inconsistent, and their combination is both accurate and ontologically valid. Two limitations shape the immediate agenda: the compositional HLI gap remains substantial and is a property of goal inference under held-out compositions, and unordered set-valued outputs are still better handled by a direct histogram baseline than by the autoregressive decoder. Future work should therefore focus on relational generalization for unseen HLI--LLI compositions, hybrid heads for set-valued anticipation, end-to-end prefix-aware encoder training, and validation on naturally recorded assistive scenarios.

\appendix
\section{Per-Seed Variance of Learned Models}

{ \begin{table*}[ht]
\caption{Per-seed variance pack for all learned models at $r=0.50$ (\%), reported as mean $\pm$ std over 3 seeds. Deterministic models (B0, B1, OntoPrior) are marked \emph{det.} and reported as means. The LLM-LoRA row is a single seed and marked \emph{1 seed}. Violation rates D1/E1/E2 are shown to two decimals given their small magnitudes; D2 is report-only. The final four rows correspond to the encoder fine-tuning ablation of Section~\ref{ssec:frozen_backbone}: R1 unfroze the last HGT layer and the fusion; R2 unfroze the full encoder; both used encoder learning rate $3\times10^{-5}$ and three seeds. EOS F1 is not reported for the fine-tuned rows because the ablation targeted next-action and HLI attribution rather than the EOS head.}
\label{tab:appendix_std}
\centering
\setlength{\tabcolsep}{2.4pt}
\adjustbox{max width=\textwidth}{%
\begin{tabular}{llcccccccc}
\toprule
Model & Split & Top-1 & Top-5 & EOS F1 & HLI top-1 & Act Jacc & LLI Jacc & E1 & E2 \\
\midrule
B0 marginal          & test  & 4.3 (det.)        & 33.8 (det.)       & 75.2 (det.)     & 11.4 (det.)       & 19.8 (det.)       & 7.8 (det.)        & 0.00 (det.)       & 0.00 (det.)       \\
B1 transition        & test  & 26.3 (det.)       & 70.3 (det.)       & 75.2 (det.)     & 11.4 (det.)       & 19.8 (det.)       & 7.8 (det.)        & 26.76 (det.)      & 0.00 (det.)       \\
B2 bag-MLP           & test  & 34.9$\pm$1.1       & 76.9$\pm$0.9       & 88.0$\pm$0.4     & 77.3$\pm$0.4       & 40.1$\pm$0.6       & 40.2$\pm$0.4       & 2.51$\pm$1.42      & 0.22$\pm$0.09      \\
B3 seq-T             & test  & 38.6$\pm$0.6       & 79.8$\pm$0.3       & 64.0$\pm$1.3     & 75.9$\pm$0.6       & 30.4$\pm$1.2       & 32.0$\pm$0.8       & 5.47$\pm$0.29      & 0.12$\pm$0.06      \\
OntoPrior            & test  & 11.2 (det.)       & 56.5 (det.)       & 75.2 (det.)     & 67.4 (det.)       & 24.4 (det.)       & 18.4 (det.)       & 0.00 (det.)       & 0.00 (det.)       \\
LLM-LoRA             & test  & 16.8 (1 seed)     & 16.8 (1 seed)     & 0.0 (1 seed)    & 30.1 (1 seed)     & 22.7 (1 seed)     & 22.1 (1 seed)     & 3.20 (1 seed)     & 2.10 (1 seed)     \\
HPD none (soft)      & test  & 38.1$\pm$0.1       & 80.4$\pm$1.2       & 56.7$\pm$1.0     & 78.2$\pm$0.5       & 22.6$\pm$1.0       & 32.1$\pm$1.0       & 5.00 (mean only)  & 0.80 (mean only)  \\
HPD full (soft)      & test  & 38.5$\pm$0.1       & 81.1$\pm$0.6       & 57.4$\pm$0.6     & 77.5$\pm$0.5       & 22.3$\pm$1.6       & 31.1$\pm$1.4       & 2.01$\pm$0.40      & 0.59$\pm$0.12      \\
HPD full+mask+setmask & test & 38.6$\pm$0.1       & 81.5$\pm$0.4       & 57.6$\pm$0.5     & 77.5$\pm$0.5       & 22.3$\pm$1.6       & 31.0$\pm$1.5       & 0.00$\pm$0.00      & 0.00$\pm$0.00      \\
\midrule
B0 marginal          & comp. & 7.3 (det.)        & 53.6 (det.)       & 69.0 (det.)     & 0.0 (det.)        & 24.4 (det.)       & 11.6 (det.)       & 0.00 (det.)       & 0.00 (det.)       \\
B1 transition        & comp. & 27.9 (det.)       & 71.1 (det.)       & 69.0 (det.)     & 0.0 (det.)        & 24.4 (det.)       & 11.6 (det.)       & 16.14 (det.)      & 0.00 (det.)       \\
B2 bag-MLP           & comp. & 25.4$\pm$0.7       & 70.9$\pm$1.4       & 85.9$\pm$0.5     & 53.1$\pm$1.6       & 25.1$\pm$0.6       & 14.0$\pm$0.5       & 1.56$\pm$1.60      & 0.03$\pm$0.02      \\
B3 seq-T             & comp. & 35.1$\pm$0.4       & 76.2$\pm$2.0       & 60.7$\pm$1.5     & 52.3$\pm$0.7       & 14.1$\pm$1.1       & 5.8$\pm$2.8        & 10.76$\pm$0.60     & 0.00$\pm$0.00      \\
OntoPrior            & comp. & 13.9 (det.)       & 58.7 (det.)       & 69.0 (det.)     & 39.4 (det.)       & 20.0 (det.)       & 14.1 (det.)       & 0.00 (det.)       & 0.00 (det.)       \\
LLM-LoRA             & comp. & 18.1 (1 seed)     & 18.1 (1 seed)     & 0.0 (1 seed)    & 25.3 (1 seed)     & 22.9 (1 seed)     & 17.8 (1 seed)     & 3.70 (1 seed)     & 2.10 (1 seed)     \\
HPD none (soft)      & comp. & 34.7$\pm$0.3       & 78.5$\pm$2.9       & 56.9$\pm$0.9     & 55.9$\pm$0.8       & 3.4$\pm$1.0        & 10.5$\pm$0.8       & 7.65 (mean only)  & 0.40 (mean only)  \\
HPD full (soft)      & comp. & 35.1$\pm$0.6       & 79.7$\pm$2.2       & 58.0$\pm$0.6     & 53.9$\pm$1.0       & 4.2$\pm$1.1        & 10.6$\pm$1.0       & 2.20$\pm$0.30      & 0.20$\pm$0.07      \\
HPD full+mask+setmask & comp.& 35.1$\pm$0.6       & 81.1$\pm$1.4       & 58.1$\pm$0.5     & 53.9$\pm$1.0       & 4.2$\pm$1.1        & 10.6$\pm$1.0       & 0.00$\pm$0.00      & 0.00$\pm$0.00      \\
\midrule
HPD full+mask+setmask, encoder FT R1 & test  & 38.7$\pm$1.1 & 81.2$\pm$1.2 & \phantom{00.0}--\phantom{$\pm$0.0} & 76.4$\pm$1.0 & 21.8$\pm$3.6 & 30.4$\pm$2.9 & 0.00$\pm$0.00 & 0.00$\pm$0.00 \\
HPD full+mask+setmask, encoder FT R2 & test  & 39.1$\pm$0.3 & 81.9$\pm$0.4 & \phantom{00.0}--\phantom{$\pm$0.0} & 77.1$\pm$0.8 & 21.4$\pm$2.3 & 31.2$\pm$1.7 & 0.00$\pm$0.00 & 0.00$\pm$0.00 \\
HPD full+mask+setmask, encoder FT R1 & comp. & 35.7$\pm$1.4 & 81.0$\pm$0.6 & \phantom{00.0}--\phantom{$\pm$0.0} & 53.1$\pm$1.2 & 4.0$\pm$1.5  & 9.4$\pm$2.4  & 0.00$\pm$0.00 & 0.00$\pm$0.00 \\
HPD full+mask+setmask, encoder FT R2 & comp. & 35.7$\pm$1.0 & 81.4$\pm$0.9 & \phantom{00.0}--\phantom{$\pm$0.0} & 53.5$\pm$1.5 & 3.2$\pm$0.6  & 10.2$\pm$1.5 & 0.00$\pm$0.00 & 0.00$\pm$0.00 \\
\bottomrule
\end{tabular}}
\end{table*}}

\section{Full Robustness Column Set}
\begin{table*}[ht]
\caption{Full column set for the robustness experiments of Section~\ref{ssec:results_robustness}, corresponding to Table~\ref{tab:robustness}. All values in \%, mean $\pm$ std over 3 seeds. E1 and E2 are exactly $0.00 \pm 0.00$ in every cell (structural guarantee of the reachability masks) and are again omitted.}
\label{tab:appendix_robustness_full}
\centering
\setlength{\tabcolsep}{2.2pt}
\adjustbox{max width=\textwidth}{%
\begin{tabular}{lllcccccc}
\toprule
Training & Condition & Split & step-1 top-1 & step-1 top-5 & step-3 top-5 & HLI top-1 & act.\ Jacc & LLI Jacc \\
\midrule
Clean       & oracle           & test  & 38.6$\pm$0.1 & 81.5$\pm$0.4 & 64.4$\pm$0.8 & 77.5$\pm$0.5 & 22.3$\pm$1.6 & 31.0$\pm$1.5 \\
Clean       & condA            & test  & 30.7$\pm$0.7 & 71.4$\pm$0.5 & 61.0$\pm$1.1 & 67.3$\pm$0.1 & 20.7$\pm$1.4 & 26.8$\pm$1.0 \\
Clean       & condB $p{=}0.10$ & test  & 35.9$\pm$0.1 & 78.5$\pm$0.7 & 63.4$\pm$0.6 & 75.4$\pm$0.2 & 21.0$\pm$1.6 & 28.3$\pm$1.3 \\
Clean       & condB $p{=}0.20$ & test  & 32.0$\pm$0.4 & 74.3$\pm$0.6 & 61.9$\pm$0.8 & 73.5$\pm$0.2 & 19.5$\pm$1.4 & 25.2$\pm$1.0 \\
Clean       & condB $p{=}0.30$ & test  & 29.9$\pm$0.5 & 71.6$\pm$0.6 & 60.7$\pm$0.3 & 71.2$\pm$0.2 & 18.4$\pm$1.3 & 22.5$\pm$1.0 \\
Noise-aware & oracle           & test  & 37.9$\pm$0.6 & 81.1$\pm$0.6 & 63.8$\pm$1.5 & 76.1$\pm$0.4 & 22.2$\pm$1.3 & 31.1$\pm$1.4 \\
Noise-aware & condA            & test  & 32.3$\pm$0.4 & 75.2$\pm$1.1 & 60.7$\pm$1.4 & 68.9$\pm$0.4 & 20.8$\pm$1.0 & 28.8$\pm$1.1 \\
Noise-aware & condB $p{=}0.10$ & test  & 35.7$\pm$0.5 & 78.8$\pm$0.3 & 62.9$\pm$0.8 & 74.5$\pm$0.2 & 21.3$\pm$1.3 & 29.3$\pm$1.1 \\
Noise-aware & condB $p{=}0.20$ & test  & 32.5$\pm$0.6 & 75.8$\pm$0.5 & 62.1$\pm$1.2 & 73.2$\pm$0.5 & 20.1$\pm$1.3 & 27.4$\pm$0.8 \\
Noise-aware & condB $p{=}0.30$ & test  & 30.8$\pm$0.6 & 73.6$\pm$0.1 & 62.0$\pm$0.2 & 71.7$\pm$0.4 & 19.2$\pm$1.1 & 25.7$\pm$0.8 \\
\midrule
Clean       & oracle           & comp. & 35.1$\pm$0.6 & 81.1$\pm$1.4 & 63.8$\pm$1.6 & 53.9$\pm$1.0 & 4.2$\pm$1.1  & 10.6$\pm$1.0 \\
Clean       & condA            & comp. & 28.9$\pm$0.6 & 72.2$\pm$1.6 & 61.3$\pm$1.7 & 43.2$\pm$0.7 & 4.1$\pm$1.0  & 9.3$\pm$0.7  \\
Clean       & condB $p{=}0.10$ & comp. & 32.0$\pm$0.3 & 76.8$\pm$1.4 & 62.9$\pm$1.2 & 52.1$\pm$1.0 & 3.9$\pm$0.9  & 9.0$\pm$0.9  \\
Clean       & condB $p{=}0.20$ & comp. & 30.2$\pm$0.6 & 73.0$\pm$1.8 & 62.5$\pm$1.2 & 49.3$\pm$0.8 & 3.5$\pm$0.8  & 7.3$\pm$0.8  \\
Clean       & condB $p{=}0.30$ & comp. & 27.0$\pm$0.5 & 69.1$\pm$1.4 & 61.2$\pm$1.6 & 47.6$\pm$0.8 & 3.2$\pm$0.8  & 6.1$\pm$0.6  \\
Noise-aware & oracle           & comp. & 35.3$\pm$0.8 & 81.6$\pm$0.3 & 62.9$\pm$1.4 & 53.3$\pm$1.2 & 5.0$\pm$2.5  & 10.8$\pm$1.3 \\
Noise-aware & condA            & comp. & 30.9$\pm$0.3 & 76.3$\pm$1.0 & 62.0$\pm$1.8 & 46.0$\pm$1.2 & 4.6$\pm$2.1  & 9.7$\pm$1.0  \\
Noise-aware & condB $p{=}0.10$ & comp. & 32.9$\pm$0.5 & 78.4$\pm$0.3 & 63.0$\pm$1.8 & 52.0$\pm$1.1 & 4.5$\pm$2.2  & 9.4$\pm$1.3  \\
Noise-aware & condB $p{=}0.20$ & comp. & 30.6$\pm$0.4 & 75.7$\pm$0.7 & 62.2$\pm$1.9 & 50.5$\pm$0.9 & 3.8$\pm$1.8  & 7.8$\pm$1.2  \\
Noise-aware & condB $p{=}0.30$ & comp. & 28.1$\pm$0.2 & 72.7$\pm$0.7 & 63.4$\pm$1.4 & 49.2$\pm$0.7 & 3.6$\pm$1.7  & 6.7$\pm$1.0  \\
\bottomrule
\end{tabular}}
\end{table*}
\section*{Declaration of Generative AI and AI-assisted Technologies in the Writing Process}

During the preparation of this work, the authors used Gemini/Claude for language refinement, grammar correction, and LaTeX formatting. The authors reviewed and edited the output as needed and take full responsibility for the content of the published article.

\bibliographystyle{elsarticle-harv}
\bibliography{references}

\end{document}